\pdfoutput=1

\documentclass[11pt]{article}

\usepackage[]{ACL2023}

\usepackage{times}
\usepackage{latexsym}

\usepackage[T1]{fontenc}

\usepackage[utf8]{inputenc}

\usepackage{microtype}

\usepackage{inconsolata}

\usepackage{algorithm}
\usepackage{algorithmic}
\usepackage{listings}
\usepackage{newfloat}
\usepackage{listings}
\usepackage{titlesec}
\usepackage{verbatim}
\usepackage{multirow}
\usepackage{graphicx}
\usepackage{booktabs}
\usepackage{amssymb}
\usepackage{amsmath}
\usepackage{bm}
\usepackage{color, soul}
\usepackage[capitalise]{cleveref}
\newcommand{\DATANAME}[0]{\textit{OASum}}
\definecolor{wingreen}{rgb}{0,0.45,0.24}
\definecolor{losered}{rgb}{1.0,0.1,0.24}

\newcommand{\win}[1]{\textcolor{wingreen}{\uparrow\hspace{0.1mm}#1} }
\newcommand{\lose}[1]{\textcolor{losered}{\downarrow\hspace{0.1mm}#1} }

\title{OASum: Large-Scale Open Domain Aspect-based Summarization}

\author{Xianjun Yang$^{1*}$ \qquad Kaiqiang Song$^{2*}$ \qquad Sangwoo Cho$^{2}$ \\ \bf
Xiaoyang Wang$^{2}$ \qquad Xiaoman Pan$^{2}$ \qquad Linda Petzold$^{1}$ \qquad Dong Yu$^{2}$\\
\texttt{\{xianjunyang,petzold\}@ucsb.edu}\\ \texttt{\{riversong,swcho,shawnxywang,xiaomanpan,dyu\}@tencent.com}\\
$^{1}$ University of California, Santa Barbara \qquad $^{2}$ Tencent AI Lab, Seattle}

\begin{document}
\maketitle
\renewcommand{\thefootnote}{\fnsymbol{footnote}}
\footnotetext[1]{Work done during Xianjun Yang's internship at Tencent AI Lab Seattle. The first two authors contributed equally.}
\renewcommand{\thefootnote}{\arabic{footnote}}

\begin{abstract}
Aspect or query-based summarization has recently caught more attention, as it can generate differentiated summaries based on users' interests.
However, the current dataset for aspect or query-based summarization either focuses on specific domains, contains relatively small-scale instances, or includes only a few aspect types.
Such limitations hinder further explorations in this direction.
In this work, we take advantage of crowd-sourcing knowledge on \texttt{Wikipedia.org} and automatically create a high-quality, large-scale open-domain aspect-based summarization dataset named \textbf{\DATANAME}, which contains more than 3.7 million instances with around 1 million different aspects on 2 million Wikipedia pages. We provide benchmark results on \DATANAME~ and demonstrate its ability for diverse aspect-based summarization generation. To overcome the data scarcity problem on specific domains, we also perform zero-shot, few-shot, and fine-tuning on seven downstream datasets. Specifically, zero/few-shot and fine-tuning results show that the model pre-trained on our corpus demonstrates a strong aspect or query-focused generation ability compared with the backbone model.
Our dataset and pre-trained checkpoints are publicly available.\footnote{\url{https://github.com/tencent-ailab/OASum}}
\end{abstract}
\section{Introduction}
Text summarization aims to provide accurate, concise, and useful information about the original inputs for users to fast browse.
Existing generic summarization or aspect agnostic summarization methods \citep{see2017get, narayan-etal-2018-dont, liu2019fine, zhang2020pegasus, liu2022brio, wang2022salience} typically generate only one summary for all different requests which is not optimal for diverse demands.
It could fail to preserve the required information that the user needs or miss important details \citep{woodsend-lapata-2012-multiple, angelidis-lapata-2018-summarizing}.
By contrast, the aspect or query-based summarization methods \citep{xu-lapata-2020-coarse, zhong-etal-2021-qmsum, ahuja-etal-2022-aspectnews} provide the flexibility of generating summaries for differentiated demands.

\begin{table*}[t]
    \centering
    \scalebox{0.90}{
    \begin{tabular}{c|c|c|c|c|c|c|c }
        \hline
        \textbf{Type} & \textbf{Dataset} & \textbf{Domain} & \#\textbf{Instances} & \#\textbf{Input Tk.} &\#\textbf{Output Tk.} & \#\textbf{Asp. Type} & \textbf{Method} \\
        \hline
        \multirow{3}{*}{Query}
        & AQualMuse  & General & 7,168 & 9,764 & 106 & 7,160 & \textbf{A} \\
        & QMSum & Meeting & 1,808 & 9,070 & 70 & 1,566 & \textbf{M} \\
        & SQuALITY & Sci-fi & 2,540 & 6,052 & 252 & 437 & \textbf{M} \\
        \hline
        \multirow{5}{*}{Aspect}
        & CovidET  & Reddit & 7,112 & 192 & 27 & 7 & \textbf{M} \\
        & MA-News  & News & 286,701 &  1,350 & 54 & 6 & \textbf{A} \\
        & NEWTS  & News & 6,000 & 602 & 74 & 50 & \textbf{M} \\
        & Wikiasp  & Wikipedia & 399,696 & 13,672 & 214 & 200 & \textbf{A} \\
         & ASPECTNews  & News & 400 & 248 & 115  & 4 & M \\ 
        \hline
        Ours & \textbf{\DATANAME}~ & Wikipedia & \textbf{3,747,569} & 1,612 & 40 & \textbf{1,045,895} & \textbf{A} \\ 
        \hline
    \end{tabular}
    }
    \caption{
    Statistics of query/aspect-based summarization datasets. The last column contains the methods of dataset creation. \textbf{A} stands for "Automatic", \textbf{M} stands for "Manual".
    \#\textbf{Input Tk.} and \#\textbf{Output Tk.} represent the number of input and output token lengths, respectively. \#\textbf{Asp. Type} is the number of all aspect types. \#\textbf{Instances} stands for the total number of $(article, summary)$ pairs in the corresponding dataset.
    }
    \label{tab:big_table}
\end{table*}

However, existing datasets for aspect-based summarization are either on a small scale \citep{ wang2022squality, bahrainian2022newts, kulkarni2020aquamuse}, only focusing on a specific domain \citep{zhong-etal-2021-qmsum, zhan2022you}, or with limited aspects \citep{frermann-klementiev-2019-inducing, hayashi2021wikiasp}. To the best of our knowledge, there is no existing dataset with millions of aspects and instances for large-scale open-domain aspect-based summarization. Models trained in a small-scale dataset with limited instances or aspects may fail to adapt to other aspects or domains in realistic open-domain scenarios.

To tackle the limitations of the existing aspect-based summarization datasets, we propose a large-scale open-domain aspect-based summarization dataset named \textbf{\DATANAME}. \cref{tab:big_table} compares \textbf{\DATANAME} with seven existing datasets for aspect or query-based summarization.
\begin{figure}[t]
\centering
\includegraphics[width=1.\linewidth]{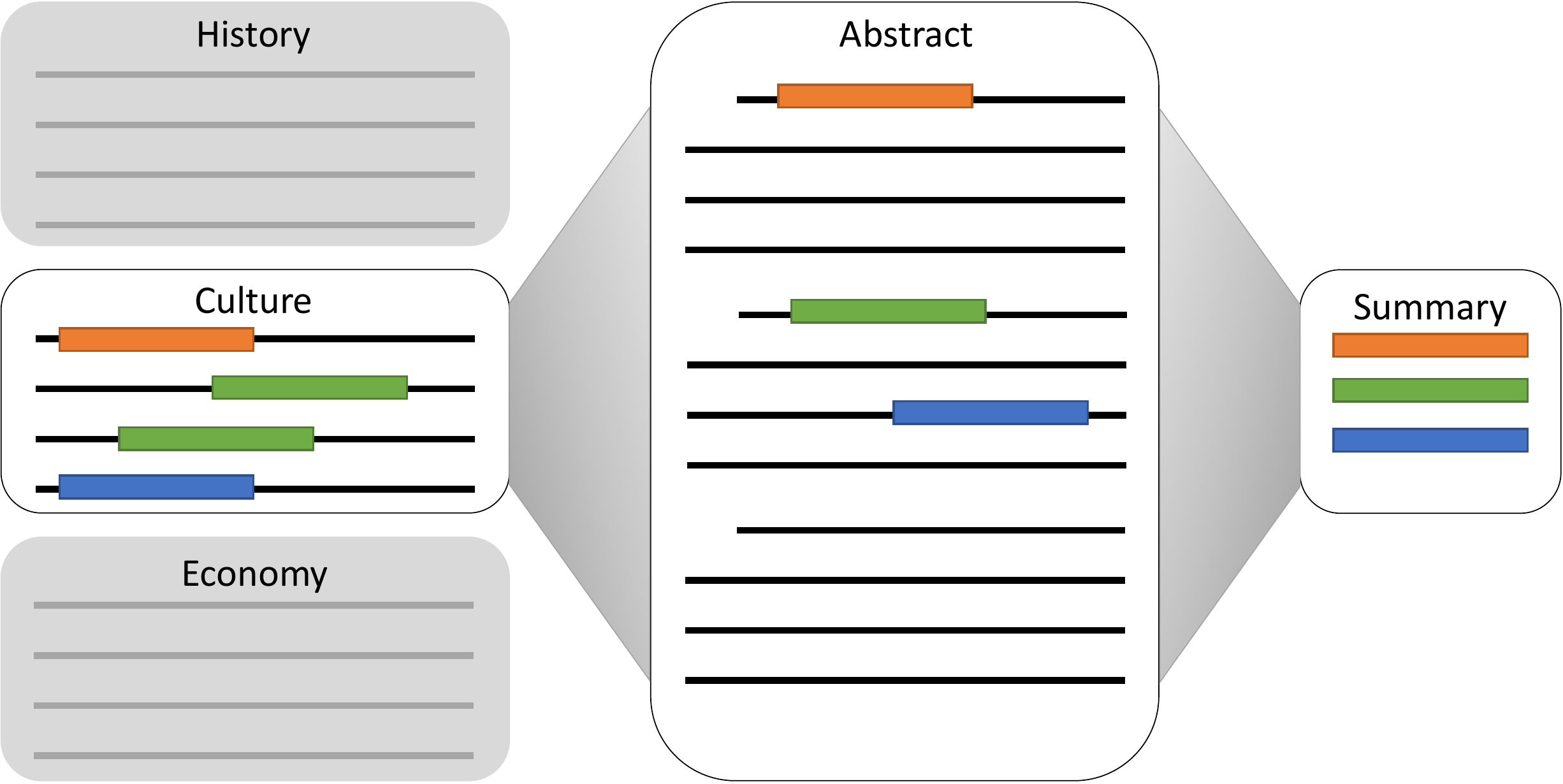}
\caption{The left section titles are naturally adopted from the Wikipedia page to serve as different aspects, while the middle abstract is the head section serving as an overall summary of the article. The right part is the corresponding aspect-based summary.}
\label{fig: OASum}
\end{figure}

To create the data, as illustrated in \cref{fig: OASum},
we take advantage of crowd-sourcing knowledge in English Wikipedia pages and parse them to collect the information on each page including the title of each section and its contents.
On the one hand, the head section is a natural abstract of each Wikipedia page.
On the other hand, the remaining sections describe different aspects of that page.
Therefore, we use the section titles as the aspect inputs and apply a rule base process to automatically select sentences in the abstract section as the matched summary for different aspects.

Specifically, we use the Wikipedia dump on 2022/06/21. It contains around 6.3 million pages after parsing.
After preprocessing, we keep approximately 2 million pages that contain around 3.7 million instances in total.
Our dataset includes 1,045,895 different aspects on 32,956 different domains (categorized with the original Wikipedia pages), providing plenty of useful information for open-domain aspect-based summarization.
It also provides abstractive summaries that are not directly extracted from the original inputs.
To ensure the quality, we perform a manual evaluation with randomly selected 66 pages, and the overall satisfaction score is 3.13 out of 5.
Based on our curated million-level aspect-based summarization corpus, we pretrain Longformer-Encoder-Decoder (LED) \citep{beltagy2020longformer} model on \textbf{\DATANAME} in an end-to-end way.
Compared with the backbone model, our pretrained model achieves better performance on six out of seven downstream tasks for the fine-tuning and zero-shot settings and all six downstream tasks for the few-shot setting.

The contributions of our work are in two folds:
\begin{itemize}
    \item
    We create the first large-scale open-domain aspect-based summarization dataset namely \textbf{\DATANAME}. 
    The statistic shows \textbf{\DATANAME} contains a variant of input lengths, highly abstractive summaries, and contents in a large number of aspects and domains.
    Overall, it contains more than 3.7M instances and 1M different aspect types.
    \item 
    We further pre-train the backbone model on \textbf{\DATANAME} and test the pretrained model with zero-shot, few-shot, and fine-tuning settings on seven downstream datasets.
    The results illustrate \textbf{\DATANAME} provides useful information that can further benefit other query/aspect-based summarization tasks. 
\end{itemize}
\section{Related Works}
\noindent \textbf{Aspect / Query based Summarization}.
Aspect-based summarization was proposed to generate summaries based on different aspects for opinions and reviews \citep{kansal2014aspect, wu2016aspect, akhtar2017aspect, angelidis-lapata-2018-summarizing, coavoux-etal-2019-unsupervised, tan-etal-2020-summarizing}.
Recent researches attempt to summarize different aspects for news~\citep{frermann-klementiev-2019-inducing, bahrainian2022newts, ahuja-etal-2022-aspectnews} and other domains~\citep{hayashi2021wikiasp, zhan2022you}.
Similarly, query-based summarization~\citep{kulkarni2020aquamuse, zhong-etal-2021-qmsum, wang2022squality} takes finer-grained questions as input for summarization.
As our \textbf{\DATANAME} contains even finer-grained aspects, we believe it can benefit both tasks.

\noindent \textbf{Wikipedia as data}.
Wikipedia has been widely used as a rich source for many NLP tasks, including Language Modeling~\citep{guo-etal-2020-wiki},
Question answering~\citep{yang2015wikiqa, rajpurkar2018know}, Information Extraction~\citep{wu2010open}, Dialogue~\citep{dinan2018wizard}, and Summarization~\citep{liu2018generating, ghalandari2020large, sun-etal-2021-document, iv-etal-2022-fruit}.
WikiAsp~\citep{hayashi2021wikiasp} directly uses external documents to generate the corresponding section contents with limited aspect types.
Comparatively, \textbf{\DATANAME} employs a matching method to obtain the aspect-based summaries from the head section of a Wikipedia page according to their similarities to the remaining page, resulting in more than one million aspect types.

\noindent \textbf{Long document summarization}.
The summarization task typically has long inputs~\citep{shen2022multi, kryscinski2021booksum, song-etal-2022-towards, cho2022toward}.
Recent Transformer-based models~\citep{radford2018improving, devlin-etal-2019-bert, lewis2020bart} with full attention require a huge amount of GPU memories during training.
Efficient transformers~\citep{beltagy2020longformer, zaheer2020big, guo-etal-2022-longt5} are proposed for handling long sequences with simplified attention.
Extract-then-generate strategies~\citep{zhong-etal-2020-extractive, pilault-etal-2020-extractive, song2020automatic, zheng2020two} have been used for such issues.
As \textbf{\DATANAME} has a large number of instances containing more than 4096 input tokens, we thus use LED~\citep{beltagy2020longformer} as our backbone model.

\section{\textbf{\DATANAME} Dataset}
\subsection{Dataset Construction}
\label{sec: data construction}
This dataset is built upon the observation that the abstract section is a natural summary for the later sections, and sentences in the abstract section may present one or more aspects described in the later sections. 
We use the English Wikipedia dump from 2022-06-20 for creating our dataset.
Originally, there are over 6.33 million pages.

\noindent \textbf{Data Cleaning}.
Each Wikipedia page is written in a special markup language.
We first adopt a tool\footnote{\url{https://github.com/panx27/wikiann}}~\cite{pan-etal-2017-cross} to remove all undesired markups (e.g., templates, internal/external links, and HTML tags) and keep section boundaries.
Next, we discard structural sections including \textit{References}, \textit{See also}, \textit{External links}, \textit{Further reading}, and \textit{Bibliography}.
We further remove structural contents such as item lists in other sections.
Finally, we split sentences using Spacy\footnote{model "en-core-web-sm", version 3.0.0}.
We collect 3.75 million non-empty pages after data cleaning.

\begin{algorithm}[h]
\caption{Greedy Mapping}
\small
\textbf{Input}: sentence $x$, set of sentences $Y$\\
\textbf{Output}: set of mapped sentences $S$

\begin{algorithmic}[1]
\STATE $S \gets \varnothing$;  // Set of mapped Sentences \\
\STATE $\textit{Score} \gets 0$; // Current ROUGE-1-Recall \\

\WHILE{$Y \setminus S \neq \varnothing$}
    \STATE $\delta \gets 0$; // Best Improvements \\
    \STATE $\eta \gets null$; // Best Candidate\\
    \FOR{$y \in Y \setminus S$}
        \STATE $S' \gets S \cup \{y\}$
        \IF{$\textit{ROUGE-1-Recall}(x, S') - \textit{Score} > \delta$}
            \STATE $\delta \gets \textit{ROGUE-1-Recall}(x, S') - \textit{Score}$;
            \STATE $\eta \gets y$;
        \ENDIF
    \ENDFOR
    \IF{$\delta \leq 0$}
        \STATE Break;
    \ENDIF
    \STATE $\textit{Score} \gets \textit{Score} + \delta$;
    \STATE $S \gets S \cup \{\eta\}$;
\ENDWHILE
\STATE return $S$
\end{algorithmic}
\label{alg: greedy mapping}
\end{algorithm}

\noindent \textbf{Aspect Summaries Construction}.
An abstract sentence should be considered as a summary sentence of the specific aspect iff it has enough content overlap with the corresponding section.
Shown in \cref{alg: greedy mapping}, we first use a greedy method to map each abstract sentence to a list of sentences in the later sections.
Then, we assign a matching score $\mathcal{S}(x, \alpha)$ for each abstract sentence $x$ and a potential aspect $\alpha$.
We use the \textit{ROUGE-1-recall} between the abstract sentence $x$ and the intersection of its mapped sentences $\mathcal{M}(x)$ and the sentences in the aspect section $Y_a$.
\begin{equation}
    \mathcal{S}(x, a) = \textit{ROUGE-1-recall}(x, Y_a \cap \mathcal{M}(x)).
    \label{eq: matching score}
\end{equation}
This score indicates the content overlap between the abstract sentence and the aspect section.
To filter out sentences with limited content overlap, an aspect-based summary includes only abstract sentences with a matching score $\mathcal{S}(x, a)$ greater or equal to a pre-defined threshold $\lambda$.
To determine the exact value of the threshold, we try $\lambda \in [0.3, 0.4, 0.5, 0.6, 0.7]$ and evaluate them manually.
Specifically, we randomly pick 66 Wikipedia pages consisting of 103 aspect-summary pairs for each threshold, and assigned them to 5 experts for evaluating the dataset quality. The Cohen's kappa between annotators is calculated to be 0.43, showing moderate agreement.
The results are shown in \cref{tab:quality}.
We then choose to use $\lambda=0.5$.
\begin{table}[h]
\centering
\scalebox{0.8}
{
\begin{tabular}{c|c|c|c|c|c}
\textbf{$\lambda=$} & \textbf{ 0.3 } & \textbf{0.4} & \textbf{0.5} & \textbf{0.6} & \textbf{0.7}\\
\toprule
{avg Score} & 2.61 & 2.85 & 3.13 & 3.05 & 2.75\\
\bottomrule
\end{tabular}
}
\caption{Summary quality with different thresholds. The scores are in the range of 1-5, representing \textit{very bad}, \textit{bad}, \textit{fair}, \textit{good}, and \textit{excellent}, respectively.}
\label{tab:quality}
\end{table}

\noindent \textbf{Data Splitting}.
We split the data into train/validation/test sets with 94\%/3\%/3\% of the Wikipedia pages after data cleaning.
After filtering out the instances where the summary is longer than the input text, we obtain 3,523,986/111,578/112,005 instances for the train/validation/test set.
In \cref{tab:seattle}, we demonstrate the aspect-based summaries constructed from the ``Seattle'' Wikipedia Page\footnote{\url{https://en.wikipedia.org/wiki/Seattle}}.
\begin{table*}[t]
    \centering
    \footnotesize
    \resizebox{.85\linewidth}{!}{
            \begin{tabular}{llll}
            \toprule
            \multicolumn{4}{p{1.\linewidth}}{ 
                {
                \textbf{History}:
                Seattle is a seaport city on the West Coast of the United States. It is the seat of King County, Washington.
                The Seattle area was inhabited by Native Americans for at least 4,000 years before the first permanent European settlers.
                Arthur A. Denny and his group of travelers, subsequently known as the Denny Party, arrived from Illinois via Portland, Oregon, on the schooner "Exact" at Alki Point on November 13, 1851.
                The settlement was moved to the eastern shore of Elliott Bay and named "Seattle" in 1852, in honor of Chief Si\'ahl of the local Duwamish and Suquamish tribes.
                Growth after World War II was partially due to the local Boeing company, which established Seattle as a center for aircraft manufacturing.
                The Seattle area developed into a technology center from the 1980s onwards with companies like Microsoft becoming established in the region; Microsoft founder Bill Gates is a Seattleite by birth.
                The stream of new software, biotechnology, and Internet companies led to an economic revival, which increased the city\'s population by almost 50,000 between 1990 and 2000.
                Seattle also has a significant musical history.}} \\
            \midrule
            \multicolumn{4}{p{1.\linewidth}}{
                {\textbf{History ; Founding}: It is the seat of King County, Washington. The Seattle area was inhabited by Native Americans for at least 4,000 years before the first permanent European settlers. Arthur A. Denny and his group of travelers, subsequently known as the Denny Party, arrived from Illinois via Portland, Oregon, on the schooner "Exact" at Alki Point on November 13, 1851. The settlement was moved to the eastern shore of Elliott Bay and named "Seattle" in 1852, in honor of Chief Si\'ahl of the local Duwamish and Suquamish tribes.}}\\
            \midrule
            \multicolumn{4}{p{1.\linewidth}}{
                {\textbf{History ; Post-war years: aircraft and software}: Growth after World War II was partially due to the local Boeing company, which established Seattle as a center for aircraft manufacturing. The stream of new software, biotechnology, and Internet companies led to an economic revival, which increased the city's population by almost 50,000 between 1990 and 2000. Seattle also has a significant musical history.}}\\
            \midrule
            \multicolumn{4}{p{1.\linewidth}}{
                {\textbf{Geography}: Seattle is situated on an isthmus between Puget Sound (an inlet of the Pacific Ocean) and Lake Washington.}}\\
            \midrule
            \multicolumn{4}{p{1.\linewidth}}{
                {\textbf{Economy}: A major gateway for trade with East Asia, Seattle is the fourth-largest port in North America in terms of container handling . Internet retailer Amazon was founded in Seattle in 1994, and major airline Alaska Airlines is based in SeaTac, Washington, serving Seattle's international airport, Seattle–Tacoma International Airport.}}\\
            \midrule
            \multicolumn{4}{p{1.\linewidth}}{
                {\textbf{Culture}: Between 1918 and 1951, nearly two dozen jazz nightclubs existed along Jackson Street, from the current Chinatown/International District to the Central District. The jazz scene nurtured the early careers of Ray Charles, Quincy Jones, Ernestine Anderson, and others. Seattle is also the birthplace of rock musician Jimi Hendrix, as well as the origin of the bands Nirvana, Pearl Jam, Soundgarden, Heart, Alice in Chains, Foo Fighters, and the alternative rock movement grunge.}}\\
            \midrule
            \multicolumn{4}{p{1.\linewidth}}{
                {\textbf{Demographics}: Today, Seattle has high populations of Native, Scandinavian, Asian American and African American people, as well as a thriving LGBT community that ranks sixth in the United States by population.}}\\ 
            \bottomrule
            \end{tabular}
    }
\caption{Example of aspect-based summaries constructed from ''Seattle''. We only show part of the aspect summaries. }
\label{tab:seattle}
\end{table*}

\subsection{Data Statistics and Analysis}
\label{sec: data properties}
In this section, we demonstrate the properties of our dataset from different perspectives including the statistics of input and output length, abstractiveness, aspect distribution, and page ontology.
\noindent \textbf{Length}.
On average, the input documents have 1,856.09 tokens or 62.23 sentences, and the output summary contains 48.61 output tokens or 1.77 sentences.
\begin{figure}[h]
\centering
\includegraphics[width=.95\linewidth]{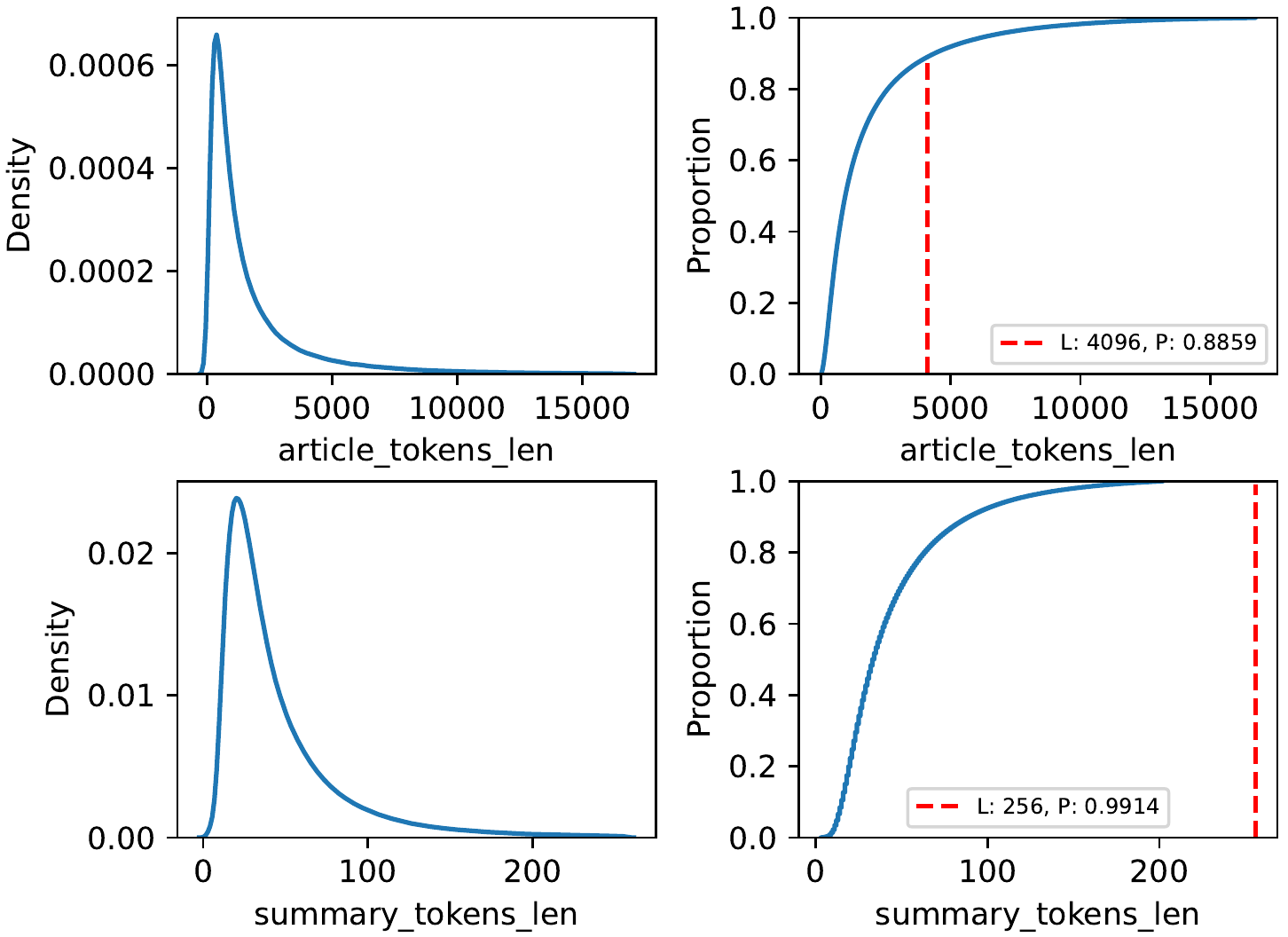}
\caption{Input (Top) and output (Bottom) length in terms of tokens with Probability Density Functions (Left) and Cumulative Distribution Functions (Right). The red dashed lines represent the truncation we used for model training. L and P represent the token length and cumulative probability, respectively.}
\label{fig: length}
\end{figure}
In \cref{fig: length}, we further plot the length distribution functions for both inputs and outputs.
We find \textbf{\DATANAME} contains a variety of lengths for both inputs and outputs.
The inputs can range from 4 tokens to 78,498 tokens, while the outputs can range from 3 to 9,792.
This creates a playground suitable for tackling long-tail problems that involve both lengthy inputs and extended summaries.
In addition, the compression ratios of \textbf{\DATANAME} are distributed widely from 0.68\footnote{ We filtered out cases in which the summary is longer than the input document in terms of words. However, this compression ratio is calculated based on its tokens.} to 32,148, which may promote the research of generating summaries with different granularity.

\noindent \textbf{Abstractiveness}.
We use novel n-gram ratios between the article and summary for measuring the abstractiveness of the summary.
More than 15.96/59.45/81.00/89.68 percent of unique 1/2/3/4-grams have not appeared in the original input.
This indicates the summary is highly abstractive.
\begin{figure}[h]
\centering
\includegraphics[width=.75\linewidth]{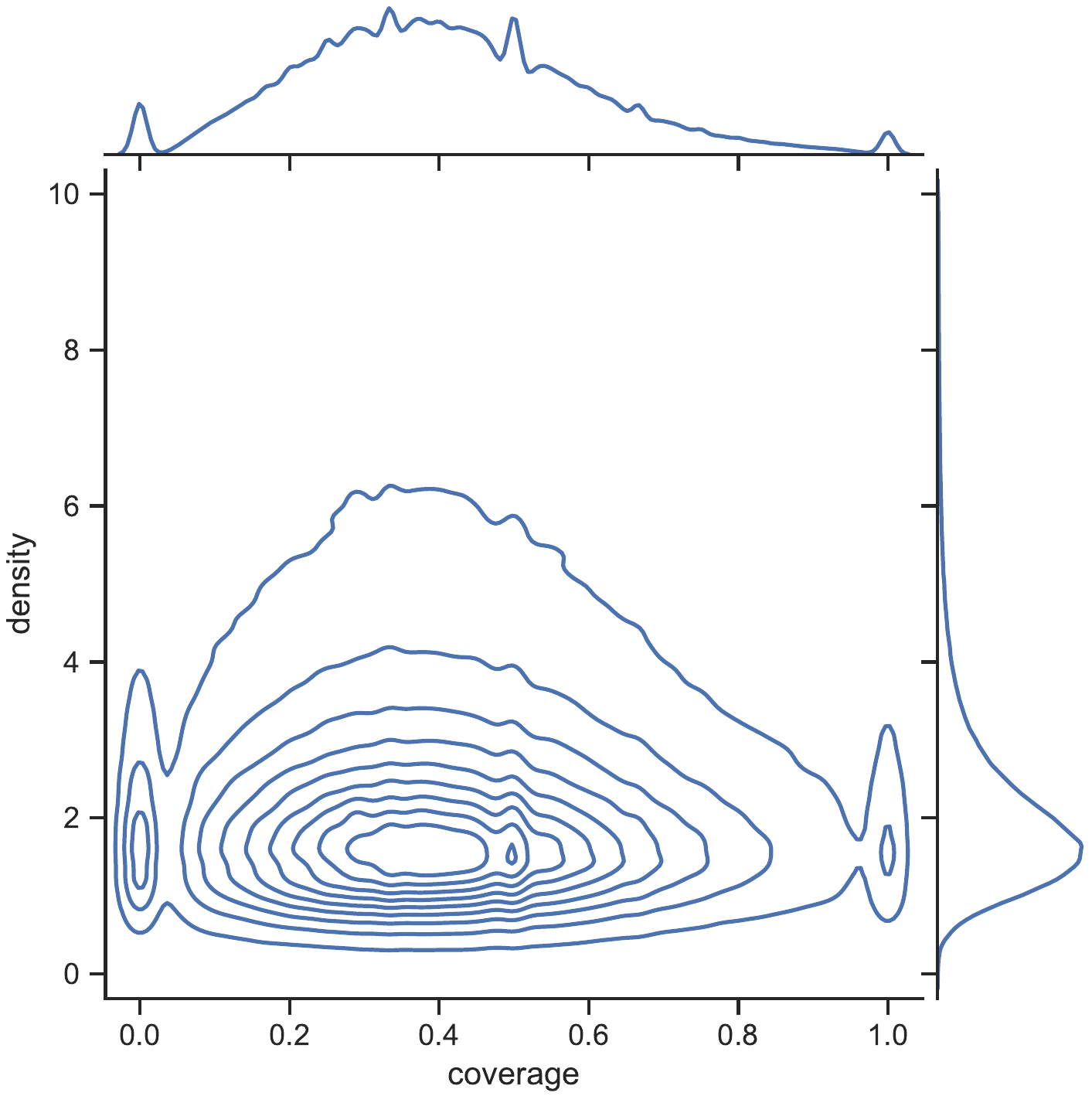}
\caption{
Normalized bi-variate density plot of bi-gram coverage vs. density for 95\% of the data.
}
\label{fig: coverage vs density}
\end{figure}
Moreover, we follow \citep{grusky-etal-2018-newsroom} and visualize the bi-variate distribution of bi-gram\footnote{We explained the reason for using bi-gram coverage and density instead of uni-gram in \cref{appendix: bi-gram}} coverage and density over \textbf{\DATANAME} in \cref{fig: coverage vs density}.
It shows that \textbf{\DATANAME} covers a large range of summarization abstractiveness styles in terms of coverage and density.

\begin{figure}[t]
\centering
\includegraphics[width=.75\linewidth]{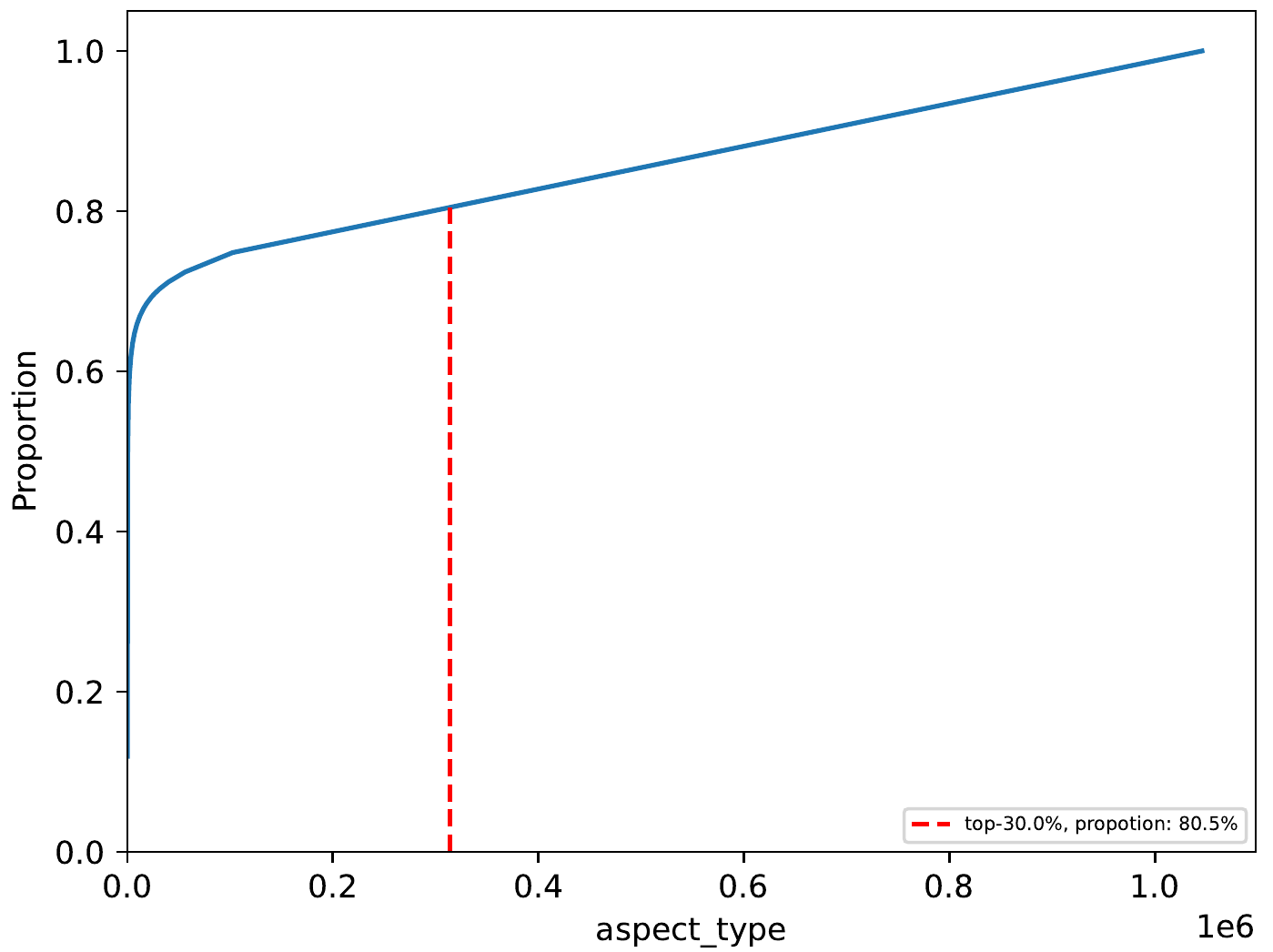}
\caption{Cumulative proportion of aspect distribution. The horizontal axis represents the sorted aspects from high frequency to low frequency.  }
\label{fig: asp_dis}
\end{figure}

\noindent \textbf{Aspects}.
In \cref{tab:big_table}, we compare \textbf{\DATANAME} with other query/aspect-based summarization datasets.
\textbf{\DATANAME} contains a significantly larger amount of aspect types.
On average, there are 1.82 aspects per article and 99\% articles have less than 9 aspects per single document. 
\begin{figure}[h]
\centering
\includegraphics[width=.75\linewidth]{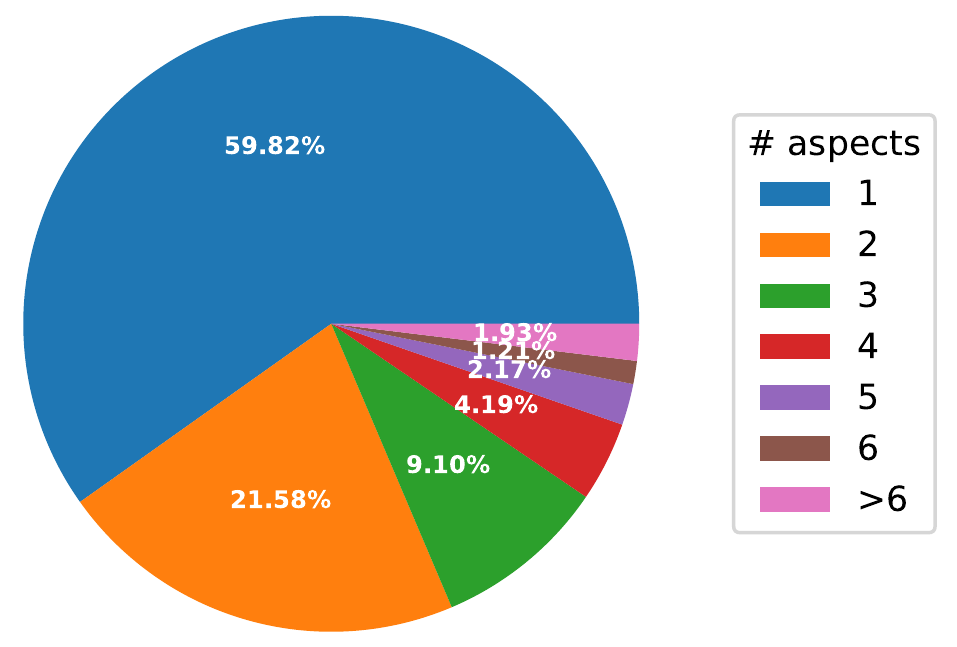}
\caption{
The pie chart for aspects per article.
}
\label{fig:aspect_dis}
\end{figure}
As shown in Fig.~\ref{fig:aspect_dis}, although 59.82\% articles only have one aspect, there are around 40\% articles that have multiple aspects ranging from 2 to more than 6. 
In total, \textbf{\DATANAME} contains 1,045,895 different types of aspects.
The top-10 common aspects are \textit{History}, \textit{Career}, \textit{Background}, \textit{Geography}, \textit{Life}, \textit{Reception}, \textit{Description}, \textit{Early life}, \textit{Demographics} and \textit{Production}, containing 447,589, 171,447, 69,266, 45,134, 43,398, 42,664, 36,199, 34,663, 34,057 and 33,424 instances, respectively.
As shown in \cref{fig: asp_dis}, we find that the top 30\% aspect types cover 80.5\% of all the cases, while the remaining 19.5\% cases come from the other 70\% aspects.
This naturally provides open-domain and diverse multiple-aspects knowledge for aspect-based summarization.

\begin{figure}[t]
\centering
\includegraphics[width=1.\linewidth]{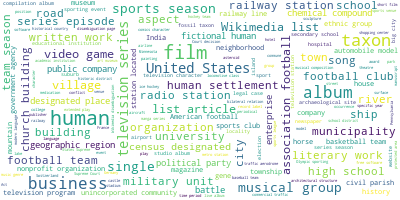} 
\caption{Word cloud based on the top 400 categories drawn from the first-level category names in \textbf{\DATANAME}. Word size is proportional to the word count. The size of the dominant category \textit{human} is reduced 10 times in corresponding to the whole category set.}
\label{fig: ontology}
\end{figure}

\noindent \textbf{Ontology}.
We analyze the domain distribution of our dataset using the ontology information provided by Wikidata's \texttt{instance of (P31)} property.
In \cref{fig: ontology}, we show the word cloud of the top 400 first-level categories of Wikipedia pages in \textbf{\DATANAME}.
In total, we cover 32,956 out of 45,042 first-level categories among Wikidata, suggesting \textbf{\DATANAME} contains text information in a large number of different domains.
To conclude, \textbf{\DATANAME} is a large-scale open-domain aspect-based summarization dataset containing varieties of input/output lengths and abstractive summaries with human-verified qualities.
\section{Baselines and Analysis}
\subsection{Metrics and Models}
In this section, we investigate the baseline models' performance over \textbf{\DATANAME}.
It includes heuristic methods(Heu), unsupervised methods, aspect-agnostic extractive methods(Ext), and aspect-based abstractive methods(Abs).
Our results are reported with ROUGE metrics~\cite{lin-2004-rouge}, including ROUGE-1, ROUGE-2, ROUGE-L, and ROUGE-Lsum.
We compare our system with extractive and abstractive summarization baselines.

\noindent \textbf{ORACLE} is generated by comparing the reference summary and each sentence in the document and obtaining the sentences with the best ROUGE scores in a greedy method~\cite{liu-lapata-2019-text}.

\noindent \textbf{RANDOM-N} Random sentences are selected for the summary. We choose the same number of sentences in the reference summary.

\noindent \textbf{LEAD-N} The leading sentences are known to be a good summary, especially in the news domain. We select the first $N$ sentences as the summary.

\noindent \textbf{SumBasic}~\cite{SumBasic07} This method takes the frequently occurring words in a document cluster for the summary.

\noindent \textbf{TextRank}~\cite{TextRank} is a graph-based approach that computes connections between sentence importance based on significant words.

\noindent \textbf{KLSum}~\cite{haghighi-vanderwende-2009-exploring} is a greedy approach to adding a sentence to the summary by minimizing KL divergence.

\noindent \textbf{LEXRANK}~\cite{LexRank} is similar to the \textit{TextRank} but tries to alleviate the redundant information by reranking selected sentences.

\noindent \textbf{Longformer-(base/large)} is a supervised extractive method. As \textbf{\DATANAME} contains long documents, we utilize the Longformer model to efficiently process the long sequence and the sentence-level Transformer layers for the sentence-level interactions. The oracle sentences are used as labels for predicting the best summary sentences.

\noindent \textbf{LED-(base/large)-\DATANAME}. We adapt LED~\cite{beltagy2020longformer} for the aspect-based summarization task.
We directly format the problem into an end-to-end sequence-to-sequence task and fine-tune the corresponding model over \textbf{\DATANAME}.
We prepend the \textit{aspect} to the input document with a \textit{[BOS]} token between them as the sequence input and use the corresponding summary as the sequence output.

\begin{table*}[t]
\small
\centering
\scalebox{1.}{
\begin{tabular}{c|c|c|cccc }
\hline
\textbf{Baselines} & \textbf{Type} & \textbf{Aspect} &  \textbf{ROUGE-1} & \textbf{ROUGE-2} & \textbf{ROUGE-L} & \textbf{ROUGE-Lsum} \\
\hline
\textbf{Oracle} & Heu & Y & 44.97 & 22.74 & 32.98 & 39.17 \\
\textit{Random-N} & Heu & N & 21.03 & 4.37 & 14.92 & 17.45 \\
\textit{LEAD-N} & Heu & N & 23.93 & 6.02 & 17.44 & 19.98 \\
\hline
\textit{SumBasic} & Ext & N & 22.79 & 5.63 & 16.55 & 19.26 \\
\textit{TexRank} & Ext & N & 23.09 & 5.90 & 15.99 & 18.62 \\
\textit{LexRank} & Ext & N & 23.95 & 6.00 & 16.81 & 19.64 \\
\textit{KLSum} & Ext & N & 22.80 & 5.59 & 15.81 & 18.40 \\
\textit{Longformer-base} (4K) & Ext & N & 30.06 & 10.75 & 22.08 & 25.35 \\
\textit{Longformer-large} (4K) & Ext & N & 30.76 & 11.21 & 22.23 & 25.78 \\
\hline
\textit{LED-base} (4K) & Abs & Y & 37.26 & 20.84 & 31.97 & 33.71 \\
\textit{LED-large} (4K) & Abs & Y & \textbf{39.61} & \textbf{22.17} & \textbf{33.34} & \textbf{35.46} \\
\hline
\end{tabular}
}
\caption{Baseline results on \textbf{\DATANAME} test set. Y and N mean including aspect or not.}
\label{tab:oasum_results_table}
\end{table*}

\subsection{Experiment Settings}
We implement our code using pytorch-lightning\footnote{\url{https://www.pytorchlightning.ai/}} and Huggingface Transformers\footnote{\url{https://github.com/huggingface/transformers}}.
The inputs and outputs are truncated to a maximum of 4096/256 tokens.
In \cref{fig: length}, the selected maximum lengths can cover 88.6\% of the entire input sequences and 99.1\% of the entire output sequences.
Since the input length is very long, we can only feed 4 instances to a single GPU for the base model and 2 instances for the large model.
For speeding up the training, Distributed Data-Parallel and Automatic Mixed Precision (FP16) are used.
Specifically, we utilize 64 NVIDIA V100 GPUs for base models and 128 NVIDIA V100 GPUs for large models for training both aspect-agnostic extractive models and aspect-based abstractive models.
The gradient accumulation step is set to 8 for reducing the communication bandwidth.
Therefore, the actual batch size is 2048.
We use Fused-Adam~\cite{kingma2015adam} implemented by NVIDIA-apex\footnote{\url{https://github.com/NVIDIA/apex}} for the optimization.
The initial learning rate is $1e-4$, and it linearly decreases to 0.
The betas are 0.9 and 0.999 respectively. 
We do not apply warm-up for \textbf{\DATANAME} training.
Weight decay is 0.01.
We evaluate the model 5 times per epoch on the validation set and pick the checkpoint with the highest average ROUGE-1/2/Lsum scores for testing.

\subsection{Results \& Analysis}
In \cref{tab:oasum_results_table}, we show the baseline model performance on \textbf{\DATANAME}.
The oracle performs the strong baseline and is used for the labels of \textit{Longformer} models. It outperforms all extractive and abstractive methods except for the ROUGE-2 and ROUGE-L of the \textit{LED-large} model. This indicates that the reference summary of \textbf{\DATANAME} is more abstractive than extractive.
The lead sentences perform similarly to the unsupervised baselines meaning that the important information is distributed to the beginning part of the documents but are not necessarily the best sentences as they under-perform the supervised methods.
Random selection is the worst choice for the summary.
For the supervised models, the extractive method outperforms the unsupervised methods but is outperformed by the abstractive methods by a large margin. We also include some generated good and bad examples as case studies in \cref{appendix: oasum_example}.
\section{Downstreams}
To verify the knowledge inside \textbf{\DATANAME} provides transfer ability, we further use the model pre-trained on \textbf{\DATANAME} for seven abstractive downstream datasets (see \cref{datasets}) including three query-based summarization datasets and four aspect-based summarization datasets, across different domains.
We test our model with zero-shot, few-shot, and fine-tuning abilities on these 7 datasets to see whether \textbf{\DATANAME} can benefit the downstream tasks.
In general, the model pre-trained on the \textbf{\DATANAME} outperforms the backbone model on 6 out of 7 tasks in the fine-tuning and zero-shot setting, 6 out of 6 tasks (w/o WikiAsp) in the few-shot setting.\footnote{WikiAsp has 20 different subsets on different domains, we only perform the results for zero-shot and fine-tuning setting.}

\begin{table}[t]
\centering
\scalebox{0.9}{
\begin{tabular}{l@{\hspace{0.8\tabcolsep}}|c|ccc}
\hline
{\textbf{Datasets}} & Models & R-1  & R-2  & R-Lsum  \\
\hline
\multirow{2}{*}{\textit{AQuaMuse}} & \textit{L} & $49.34$ & $33.26$ & $46.42$ \\
& \textit{O} & $\textbf{49.98} $ & $\textbf{34.12} $ & $\textbf{47.09} $ \\
\hline

\multirow{2}{*}{\textit{CovidET}} & \textit{L} & $\textbf{26.19}$ &  $\textbf{6.85}$ &  $\textbf{20.82}$ \\ & \textit{O} & $25.61 $ & $6.58 $ & $20.45 $ \\
\hline

\multirow{2}{*}{\textit{MA-News}} & \textit{L} & $37.8$ & $\textbf{17.43}$ & $35.3$ \\
& \textit{O} & $\textbf{38.12} $ & $17.41 $ & $\textbf{35.51} $ \\
\hline

\multirow{2}{*}{\textit{NEWTS}} & \textit{L} & $31.96$ & $10.75$ & $28.72$ \\
& \textit{O} & $\textbf{32.45} $ & $\textbf{11.64} $ & $\textbf{29.14} $ \\
\hline

\multirow{2}{*}{\textit{QMSum}} & \textit{L} & $29.52$ & $7.00$ & $25.68$ \\
& \textit{O} & $\textbf{30.30}  $ & $\textbf{7.56}  $ & $\textbf{26.67}  $ \\
\hline

\multirow{2}{*}{\textit{SQuaLITY}} & \textit{L} & $36.78$ & $8.31$ & $34.47$ \\
& \textit{O} & $\textbf{37.6} $ & $\textbf{8.81} $ & $\textbf{35.14} $ \\
\hline

\multirow{2}{*}{\textit{Wikiasp}} & \textit{L} & $22.18$ & $8.21$ & $20.48$ \\
& \textit{O} & $\textbf{22.69} $ & $\textbf{8.29} $ & $\textbf{20.92} $ \\
\hline
\end{tabular}
}
\caption{
Fine-tuning results on downstream tasks.
Wikiasp results are the average number of all 20 domains.
\textit{L} represents LED-base, \textit{O} represents LED-OASum-base.
}
\label{tab:finetune_table}
\end{table}

\begin{table*}[t]
\centering
\scalebox{0.70}{
\begin{tabular}{l@{\hspace{0.99\tabcolsep}} | c|ccc | ccc | ccc }
\hline
\multirow{2}{*}{\textbf{Datasets}} & \multirow{2}{*}{\textbf{Models}}
  & \multicolumn{3}{c|}{\textbf{Few-shot 0.3\%}} & \multicolumn{3}{c|}{\textbf{Few-shot 1\%}} & 
  \multicolumn{3}{c}{\textbf{Few-shot 3\%}} \\
 \cline{3-11}
  & & R-1 & R-2 & R-Lsum & R-1 & R-2 & R-Lsum & R-1 & R-2 & R-Lsum  \\
\hline
\multirow{2}{*}{\textit{AQuaMuse}} & \textit{L} &$32.44_{ \pm 0.91}$ & $13.89_{ \pm 0.95}$  & $29.41_{ \pm 0.91}$ & $35.29_{ \pm 1.39}$ & $16.91_{ \pm 1.30}$ & $32.16_{ \pm 1.17}$ & $37.55_{ \pm 0.57}$ & $19.88_{ \pm 0.83}$ & $34.40_{ \pm 0.71}$ \\
& \textit{O} & $38.77_{ \pm 0.53}^{\win{6.33}}$ & $20.61_{ \pm 0.80}^{\win{6.72}} $ & $35.61_{ \pm 0.70}^{\win{6.20}}$ & $40.63_{ \pm 0.20}^{\win{5.34}}$ & $22.81_{ \pm 0.80}^{\win{5.90}}$ & $37.50_{ \pm 0.39}^{\win{5.34}}$ & $41.50_{ \pm 1.04}^{\win{3.95}}$ & $24.25_{ \pm 0.92}^{\win{4.37}}$ & $38.55_{ \pm 1.01}^{\win{4.15}}$ \\
\midrule

\multirow{2}{*}{\textit{CovidET}} & \textit{L} &$20.33_{ \pm 0.01}$ & $3.75_{ \pm 0.17}$  & $16.53_{ \pm 0.15}$ & $21.19_{ \pm 0.30}$ & $4.40_{ \pm 0.02}$ & $17.39_{ \pm 0.10}$ & $22.56_{ \pm 0.18}$ & $5.07_{ \pm 0.07}$ & $18.37_{ \pm 0.15}$ \\
& \textit{O} & $22.00_{ \pm 0.12}^{\win{1.67}}$ & $4.58_{ \pm 0.14}^{\win{0.83}}$ & $17.90_{ \pm 0.12}^{\win{1.37}}$ & $22.16_{ \pm 0.02}^{\win{0.97}}$ & $4.58_{ \pm 0.02}^{\win{0.18}}$ & $18.02_{ \pm 0.03}^{\win{0.64}}$ & $22.73_{ \pm 0.34}^{\win{0.17}}$ & $5.02_{ \pm 0.15}^{\lose{0.05}}$ & $18.40_{ \pm 0.24}^{\win{0.03}}$ \\
\midrule
	
\multirow{2}{*}{\textit{MA-News *}} & \textit{L} &$20.12_{ \pm 0.03}$ & $5.08_{ \pm 0.01}$ & $18.52_{ \pm 0.03}$ & $20.41_{ \pm 0.05}$ & $5.40_{ \pm 0.02}$ & $18.84_{ \pm 0.03}$ & $22.07_{ \pm 0.02}$ & $6.63_{ \pm 0.02}$ & $20.41_{ \pm 0.01}$ \\
& \textit{O} & $24.15_{ \pm 0.17}^{\win{4.03}}$ & $7.37_{ \pm 0.05}^{\win{2.28}}$ & $22.22_{ \pm 0.13}^{\win{3.69}}$ & $25.12_{ \pm 0.01}^{\win{4.71}}$ & $7.98_{ \pm 0.01}^{\win{2.58}}$ & $23.11_{ \pm 0.01}^{\win{4.27}}$ & $27.58_{ \pm 0.08}^{\win{5.51}}$ & $9.67_{ \pm 0.02}^{\win{3.03}}$ & $25.49_{ \pm 0.07}^{\win{5.08}}$ \\
\midrule

\multirow{2}{*}{\textit{NEWTS}} & \textit{L} &$26.24_{ \pm 0.03}$ & $7.35_{ \pm 0.11}$  & $23.47_{ \pm 0.05}$ & $26.77_{\pm 0.53}$ & $8.16_{ \pm 0.23}$ & $24.63_{ \pm 0.66}$ & $27.92_{ \pm 0.02}$ & $8.47_{ \pm 0.28}$ & $25.06_{ \pm 0.04}$ \\
& \textit{O} & $27.75_{ \pm 0.49}^{\win{1.50}}$ & $8.10_{ \pm 0.02}^{\win{0.75}}$ & $24.77_{ \pm 0.40}^{\win{1.30}}$ & $28.59_{ \pm 0.09}^{\win{1.82}}$ & $8.66_{ \pm 0.03}^{\win{0.50}}$ & $25.50_{ \pm 0.06}^{\win{0.88}}$ & $28.15_{ \pm 0.67}^{\win{0.24}}$ & $8.80_{ \pm 0.02}^{\win{0.33}}$ & $25.27_{ \pm 0.54}^{\win{0.20}}$ \\
\midrule

\multirow{2}{*}{\textit{QMSum}} & \textit{L} & $19.80_{ \pm 0.63}$ & $3.23_{ \pm 0.01}$  & $17.28_{ \pm 0.14}$ & $22.58_{ \pm 3.58} $ & $3.80 _{\pm 0.31}$ & $19.70_{ \pm 2.08}$ & $24.52_{ \pm 0.70}$ & $4.64_{ \pm 0.38}$ & $21.39_{ \pm 0.50}$ \\
& \textit{O} & $22.98_{ \pm 2.06}^{\win{3.17}}$ & $4.40_{ \pm 0.29}^{\win{1.17}}$ & $19.88_{ \pm 0.87}^{\win{2.60}}$ & $24.51_{ \pm 1.88}^{\win{1.93}}$ & $4.53_{ \pm 0.26}^{\win{0.73}}$ & $21.38_{ \pm 1.06}^{\win{1.68}} $ & $25.48_{ \pm 0.01}^{\win{0.96}}$ & $5.30_{ \pm 0.16}^{\win{0.67}}$ & $22.21_{ \pm 0.05}^{\win{0.82}}$ \\
\midrule

\multirow{2}{*}{\textit{SQuaLITY}} & \textit{L} &$26.27_{ \pm 0.68}$ & $4.39_{ \pm 0.01}$  & $24.72_{ \pm 0.69}$ & $26.79_{ \pm 1.24}$ & $4.58_{ \pm 0.14}$ & $25.27_{ \pm 1.16}$ & $31.52_{ \pm 0.66}$ & $ 5.79_{ \pm 0.22} $ & $29.55_{ \pm 0.63}$ \\
& \textit{O} & $29.05_{ \pm 0.23}^{\win{2.78}}$ & $5.19_{ \pm 0.05}^{\win{0.80}}$ & $27.18_{ \pm 0.28}^{\win{2.45}}$ & $30.72_{ \pm 0.20}^{\win{3.93}}$ & $5.75_{ \pm 0.06}^{\win{1.17}}$ & $28.80_{ \pm 0.20}^{\win{3.53}}$ & $33.05_{ \pm 0.68}^{\win{1.53}}$ & $6.71_{ \pm 0.01}^{\win{0.92}}$ & $31.04_{ \pm 0.63}^{\win{1.49}}$ \\
\midrule

\end{tabular}
}
\caption{
Few-shot performance. MA-News results are under 0.03\%, 0.1\%, and 0.3\%.
\textit{L} represents LED-base, \textit{O} represents LED-OASum-base.
}
\label{tab:few-shot}
\end{table*}
\begin{table}[t]
\centering
\scalebox{0.90}{
\begin{tabular}{l@{\hspace{0.8\tabcolsep}}|c|ccc}
\hline
{\textbf{Datasets}} & Models & R-1  & R-2  & R-Lsum  \\
\hline
\multirow{2}{*}{\textit{AQuaMuse}} & \textit{L} & $24.98$ & $9.22$ & $22.93$ \\
& \textit{O} & $\textbf{36.80}$ & $\textbf{18.18}$ & $\textbf{33.50}$ \\
\hline	
		
\multirow{2}{*}{\textit{CovidET}} & \textit{L} & $ 14.61 $ &  $ \textbf{3.08} $ &  $ 12.37$ \\ 
& \textit{O} & $\textbf{15.75} $ & $ 2.01 $ & $ \textbf{12.72} $ \\
\hline
		
\multirow{2}{*}{\textit{MA-News}} & \textit{L} & $17.01$ & $5.56$ & $15.82$ \\
& \textit{O} & $\textbf{20.06}$ & $\textbf{5.81}$ & $\textbf{18.41}$ \\
\hline

\multirow{2}{*}{\textit{NEWTS}} & \textit{L} & $\textbf{26.71}$ & $\textbf{8.49}$ & $\textbf{22.14}$ \\
& \textit{O} & $24.06$ & $6.91$ & $21.04$ \\
\hline

\multirow{2}{*}{\textit{QMSum}} & \textit{L} & $13.96$ & $2.29$ & $12.70$ \\
& \textit{O} & $\textbf{22.51}$ & $\textbf{3.27}$ & $\textbf{19.95}$ \\
\hline
			
\multirow{2}{*}{\textit{SQuaLITY}} & \textit{L} & $26.87$ & $3.69$ & $25.41$ \\
& \textit{O} & $\textbf{30.54}$ & $\textbf{5.72}$ & $\textbf{28.86}$ \\
\hline
\multirow{2}{*}{\textit{Wikiasp}} & \textit{L} & $8.90$ & $1.06$ & $8.04$ \\
& \textit{O} & $\textbf{15.61}$ & $\textbf{2.75}$ & $\textbf{13.91}$ \\
\hline
\end{tabular}
}
\caption{Zero-shot results on downstream tasks. Wikiasp results are the average number on all 20 domains. \textit{L} represents LED-base, \textit{O} represents LED-OASum-base.} 
\label{tab:zero_shot_table}
\end{table}

\subsection{Experiment Settings}
For all downstream tasks, we only test the base model to demonstrate the ability of our pretrained checkpoint in an end-to-end setting.
We experiment with different decoding hyper-parameters and find the $length\_penalty = 1.0$, $num\_beams = 4$, and $no\_repeat\_ngram_size = 3$ consistently achieve optimal performance on multiple datasets in the zero-shot setting. Thus, we keep these parameters for all downstream task experiments.
For the backbone \textbf{LED-base} model (denoted as \textbf{L} ), we initialize the model using the checkpoint provided by \cite{beltagy2020longformer} on Huggingface\footnote{\url{https://huggingface.co/allenai/led-base-16384}}.
On top of the backbone model, our model checkpoint is further fine-tuned on \textbf{LED-\DATANAME} (denoted as \textbf{O} ) for 20 epochs.
Notice that for fine-tuning and zero-shot scenarios, the Wikiasp results are reported on an average of 20 domains tested independently.

\subsection{Fine-tuning Settings}
For fine-tuning experiments, we directly fine-tune the model on the whole training set and report the ROUGE scores on the test set by selecting the best-performing checkpoint on the validation set.
We present all the fine-tuning results in \cref{tab:finetune_table} with ROUGE-1, ROUGE-2, and ROUGE-Lsum scores.
In general, models fine-tuned on our checkpoint consistently perform better and demonstrate a strong advantage in ROUGE scores. 
\cref{appendix: wikiasp_full_results} shows the complete results of the 20 domains of Wikiasp.
We find that our fine-tuned models outperform the backbone model on most of the domains, with only a few exceptions.
Overall, our experiments demonstrate that fine-tuning the backbone model on \textbf{\DATANAME} is an effective approach for improving performance on a variety of aspects or query-based summarization tasks. 

\subsection{Few-Shot Settings}
For few-shot experiments, we randomly pick 0.3\%, 1\%, and 3\% of the training data, then perform 60-epoch training on the picked low-resource samples.
To compensate randomness, we conduct all experiments for three times using different random seeds to pick the training data.
The results are reported based on the average and variance of ROUGE scores over experiments with different random seeds. 
\cref{tab:few-shot} includes the few-shot performance of the backbone model and our model. 
An obvious superiority is demonstrated based on our checkpoint models for almost all R-1, R-2, and R-Lsum scores under 0.3\%, 1\%, and 3\% settings for every aspect or query-based summarization dataset.
For all the datasets, we can achieve substantial advancements of around 1 to 7 points improvements under ROUGE evaluation. 
Besides, the greatest improvements almost always happen in extremely low-resource($0.3\%$) scenarios, demonstrating the great adaptability of our model for various domains. Given the difficulty of gathering such data, we think our findings are beneficial across many disciplines.
In \cref{tab:case-led-oasum2}, we also show some typical examples.

\subsection{Zero-shot Settings}
For the zero-shot experiments, we only test the models on the whole test set without any optimization of the training data.
The zero-shot evaluation results are demonstrated in \cref{tab:zero_shot_table}.
The complete results on 20 domains of Wikiasp are also shown in \cref{tab:wikiasp_full_table}.
As we can see, except for NEWTS datasets, our LED-\DATANAME~ consistently achieves significantly better results in almost all evaluation metrics.
We believe this improvement comes from the rich knowledge contained in the large corpus learned during the pre-training.
The performance almost doubles on Wikiasp and AQuaMuse, validating that the knowledge is successfully transferred into the generation process. More case studies can be found in \cref{tab:case-led-oasum} and \cref{tab:case-led-oasum2}.

\section{Conclusions}
In summary, we contribute the first large-scale open-domain aspect-based summarization corpus collected using Wikipedia section titles as aspects by rules with good quality. Detailed statistics reveal many different aspects of the corpus, confirming its broader coverage.
We also outline the methods we use for pre-training the generative language models and present abstractive and extractive results as a baseline for future work.
Furthermore, we prove that our pre-trained model can consistently improve seven widely-used downstream tasks, especially in few-shot and zero-shot settings. 
We hope our data and pre-trained models can further foster relevant research in this area.

\section{Limitations}
First of all, our \DATANAME~ inevitably contains inappropriate summaries not strongly correlated with certain aspects since it is automatically curated. The model trained on it could furthermore hold such misinformation and affect other downstream tasks. But we hope the large-scale training can alleviate such effects to a minimum. At the current stage, we are not responsible for any products directly built on our results. In the future, a potential denoising mechanism could be designed to further reduce the noisy summaries.

Secondly, we only opt for end-to-end extraction, which requires large computational memory and cost that may not be afforded by everyone. Thus, a meaningful direction would be investigating other extract-then-summarize two-step methods for dealing with long document summarization. Besides, our vanilla dataset contains millions of summaries that are difficult for certain researchers with limited computational resources to directly reproduce results on. We recommend using a small subset of our corpus if enough computational capability is not immediately available.

Finally, we only explore a simple strategy for controlling the summarization based on input aspects. However, we find it can not always guarantee aspect-focused generation. How to efficiently and accurately generate specific summaries by confining aspects is not only challenging for model design but also difficult for humans to evaluate. We leave these issues for future work.

\section{Acknowledgments}
This work was done when Xianjun Yang was doing an internship at Tencent AI Lab Seattle. Xianjun Yang was supported in part by the UC Santa Barbara NSF Quantum Foundry funded via the Q-AMASEi program under NSF award DMR1906325.

\bibliography{anthology,custom}
\bibliographystyle{acl_natbib}

\appendix

\section{Details in Data Statistics}
\label{appendix: data}

\subsection{Top 50 Aspects}
\label{appendix: top_aspects}
\begin{table*}
\centering
\scalebox{.85}
{
\begin{tabular}{c|c|c|c|c|c }
\textbf{Aspect} & \textbf{ Count } & \textbf{Aspect} & \textbf{ Count } & \textbf{Aspect} & \textbf{ Count } \\
\toprule
 
\textit{History}  & 447,589 & \textit{Career}  & 171,447 & \textit{Background} & 69,266 \\
\textit{Geography}  & 45,134  & \textit{Life}  & 43,398 & \textit{Reception}  &  42,664 \\
\textit{Description}  & 36,199 & \textit{Early life}  & 34,663  & \textit{Demographics}  & 34,057 \\
\textit{Production}  & 33,424  & \textit{Plot}  & 32,331 & \textit{Overview}  & 23,465 \\
\textit{Professional career} & 21,237 & \textit{Political career}  & 20,232 & \textit{Club career}  & 18,520 \\
\textit{Release}  & 17,867 & \textit{Playing career}  & 17,735 & \textit{Life and career}  & 17,672  \\
\textit{Personal life}  & 15,786  & \textit{Development}  & 14,253 & \textit{Early life and education}  & 13,056 \\
\textit{Critical reception}  & 12,889 & \textit{Track listing}  & 11,176  & \textit{Route description}  & 10,065  \\
\textit{Legacy}  & 9,982 & \textit{International career}  & 9,582 &  \textit{Gameplay}  & 8,533 \\
\textit{Location}  &  8,379 & \textit{Coaching career}  &7,860  & \textit{Aftermath}   & 7,672  \\
\textit{Taxonomy }  & 7,570  & \textit{College career }   &  7,302  & \textit{Synopsis }  & 7,063  \\
\textit{Design }   &  6,651  & \textit{ Demographics ; 2010 census}  & 6,541  & \textit{ Education}   &  6,461  \\
\textit{Distribution and habitat }  & 6,344  & \textit{ Early life and career}   & 6,328  &
\textit{Description and history }  & 5,711  \\
\textit{ Death}   &  5,709 & \textit{Early years }  & 5,664  & \textit{Awards }   &  5,657  \\
\textit{Structure }  & 5,541   & \textit{ Composition}   & 5,535  &
\textit{Music video }  & 5,513  \\
\textit{Politics }   & 5,191  & \textit{Function }  & 5,061  & \textit{Distribution }   & 5,034   \\
\textit{ Origins}  & 4,942  & \textit{ Publication history}   &  4,809 & &   \\

\bottomrule
\end{tabular}
}
\caption{The most common 50 aspects and their frequencies.}
\label{tab:top-50-aspect}
\end{table*}

In \cref{tab:top-50-aspect}, We show the most common 50 aspects in \textbf{\DATANAME} and their frequencies. As we can see, those aspects naturally cover many perspectives of an article, serving as good and diverse aspects to be summarized with.

\subsection{Top 50 Categories}
\label{appendix: top_categories}
\begin{table*}
\centering
\scalebox{1.0}
{
\begin{tabular}{c|c|c|c|c|c }
\textbf{Wikidata ID} & \textbf{Category} & \textbf{ Count } & \textbf{Wikidata ID} & \textbf{Category} & \textbf{ Count } \\
\toprule
Q5 & \textit{ human } & $572,975$ &Q11424 & \textit{ film } & $45,427$  \\
Q16521 &\textit{ taxon } & $40,182$ & Q482994 &	\textit{ album } & $35,055 $ \\
Q4830453 &\textit{ business } & $33,406$ & Q134556 &	\textit{ single } & $21,768$  \\
Q215380 &\textit{ musical group } & $20,615$ & Q27020041	&\textit{ sports season } & $18,045$  \\
Q13406463 &\textit{ Wikimedia list article } & $17,150$ & Q7889&	\textit{ video game } & $14,776$  \\
Q486972&\textit{ human settlement } & $14,744$ & Q7725634	&\textit{ literary work } & $14,710$  \\
Q5398426&\textit{ television series } & $13,969$ & Q34442	&\textit{ road } & $13,224$  \\
Q43229 &\textit{ organization } & $13,085$ & Q7366	&\textit{ song } & $12,516$  \\
Q55488 &\textit{ railway station } & $11,662$ & Q476028 &\textit{ association. } & $10,652$  \\
Q14350 &\textit{ radio station } & $10,120$ & Q532 &\textit{ village } & $9,794$  \\
Q9826 &\textit{ high school } & $9,212$ & Q11446 	&\textit{ ship } & $9,060$  \\
Q1093829 &\textit{ city. } & $8,403$ & Q16970	&\textit{ church building } & $8,376$  \\
Q176799 &\textit{ military unit } & $7,845$ & Q47461344	&\textit{ written work } & $6,917$  \\
Q21191270 &\textit{ television. } & $6,870$ & Q41176	&\textit{ building } & $6,543$  \\
Q4022 &\textit{ river } & $6,502$ & Q498162	&\textit{ census. } & $6,402$  \\
Q3918 &\textit{ university } & $5,833$ & Q3914	&\textit{ school } & $5,769$  \\
Q15127012 &\textit{ town. } & $5,674$ & Q3957	&\textit{ town } & $5,576 $ \\
Q6881511 &\textit{ enterprise } & $5,542$ & Q15632617	&\textit{ fictional human } & $5,502 $ \\
Q11173 &\textit{ chemical compound } & $5,404$ & Q7278	&\textit{ political party } & $5,284$  \\
Q178561 &\textit{ battle } & $5,159$ & Q891723	&\textit{ public company } & $4,916$  \\
Q17343829 &\textit{ unincorporated.} & $4,792$ & Q1115575	&\textit{ civil parish } & $4,672 $ \\
Q163740 &\textit{ nonprofit organization } & $4,418$ & Q123705	&\textit{ neighborhood } & $4,413$  \\
Q515 &\textit{ city } & $3,900$ & Q15416	&\textit{ television program } & $3,864$  \\
Q3231690 &\textit{ automobile model } & $3,811$ & Q41710	&\textit{ ethnic group } & $3,747$  \\
Q7187 &\textit{ gene } & $3,724$ & Q74817647	&\textit{ aspect. } & $3,719$  \\

\bottomrule
\end{tabular}
}
\caption{The most common 50 categories, the corresponding Wikidata IDs, and their frequencies. \textit{unincorporated.}, \textit{aspect.}, \textit{city.}, \textit{town.}, \textit{association.}, \textit{television.} and \textit{census.} are short for \textit{unincorporated community in the United States}, \textit{aspect in a geographic region}, \textit{city/town of the United States}, \textit{association football club}, \textit{television series episode}, \textit{census-designated place}, respectively. }
\label{tab:top-50-category}
\end{table*}

In \cref{tab:top-50-category}, We show the most common 50 categories of Wikipedia pages in \textbf{\DATANAME} and their frequencies. In general, the top-50 and top-10\% categories take up around 57.84\%, and 93.51\% of all the categories, respectively.

\subsection{Bi-gram coverage and density}
\label{appendix: bi-gram}
We notice that uni-gram coverage and density presented in the \cite{grusky-etal-2018-newsroom} could only represent the token level extractiveness.
However, summarizers typically extract self-contained~\cite{cho-etal-2020-better} text spans to construct a summary.
It usually works on sentence-level or sub-sentence level.
In such cases, the token-level extractiveness cannot well represent how extractiveness the instance is.
It becomes worse when the input document is long enough, containing different pieces of summary tokens in different places of the document.
On the country, bi-gram coverage and density reduce the chance of wrongly representing the extractiveness of the instances.
Thus, in this work, we choose to use bi-gram coverage and density for presenting the extractiveness / abstractiveness of instances.

\section{Details in Experiments}
\label{appendix: experiments}

\subsection{Datasets}
\label{datasets}
We list the 7 downstream datasets below, their statistics are shown in \cref{tab:big_table}:

\noindent \textbf{AQuaMuse}~\citep{kulkarni2020aquamuse} is a Query-based multi-document summarization(qMDS) dataset built by automatically mining qMDS examples from question-answering datasets and large document corpora. We follow the preprocessing steps in \citep{vig-etal-2022-exploring} to build the AQuaMuse based on Version 3 and get a train/validation/test split of 5,784/637/747. For multiple documents, we directly concatenate them together as inputs in a natural order.

\noindent \textbf{CovidET}~\citep{zhan2022you} includes abstractive summaries of seven emotion triggers related to COVID-19 Reddit posts written by humans.
Following their public repository\footnote{\url{https://github.com/honglizhan/CovidET}}, we successfully build 4,419/1,077/1,616 instances for train/validation/test. Notice that in their dataset, one instance may have several different reference summaries. We follow their evaluation considering the average ROUGE scores if multiple references exist.

\noindent \textbf{MA-News}~\citep{frermann-klementiev-2019-inducing} synthesize multi-aspect summaries by interleaving paragraphs of $n_d$ documents belonging to different aspects and pairing the document with each of its $n_d$ components’ reference summaries.
It includes 284,700/14,589/12,800 train/validation/test summarization pairs.

\noindent \textbf{NEWTS}~\citep{bahrainian-etal-2022-newts} contains 4800 training and 1,200 testing aspect-based abstractive summaries annotated by humans derived from the well-known CNN/Dailymail \citep{hermann2015teaching, nallapati2016abstractive} dataset. Each article contains two general aspects, such as economics and politics.
We randomly split the original 1,200 testing samples into 300 instances for validation and 900 for the test.
 
\noindent \textbf{QMSum}~\citep{zhong-etal-2021-qmsum} select and summarize relevant spans of meetings in response to a specific query. It contains 1,257, 272, and 279 training, validation, and test instances, respectively. The query is usually a general question such as \textit{summarize the whole meeting .} or a specific query like \textit{how did marketing design the product evaluation ?}.

\noindent \textbf{SQuALITY} \citep{wang2022squality} is a dataset of question-focused long-document summaries built on the public-domain short stories by hiring highly-qualified contractors to read stories and write original summaries from scratch.
Documents are an average of 5,199.4 tokens long, while responses and plot summaries are 237.1, and 441.9 tokens long on average, respectively.

\noindent \textbf{Wikiasp} \citep{hayashi2021wikiasp} provides multi-domain aspect-based summarization by using the section titles and boundaries of each Wikipedia article for aspect annotation and all available references as source with an average length of 13,672.
It contains 20 different domains and 200 aspects, we present the averaged results on all 20 domains.

\subsection{Hyper-parameters}
\label{appendix: hyper-parameters}
\noindent \textbf{Fine-tuning}.
For downstream tasks, we fine-tune the model with 20 epochs on WikiAsp and 50 epochs on the remaining datasets.
We then pick the checkpoint with the best validation average ROUGE performance to test its final performance on the testing data.
In \cref{tab:hypama_finetune}, we show the hyper-parameters used in the fine-tuning setting of different datasets.
For decoding, we keep no\_repeat\_ngrams as 3, the beam size is set to 4, and the length penalty is set to 1.0.
We use a linearly decreasing learning rate schedule on all tasks without any warm-up.
The weight decay is set to 0.01.

\begin{table}[h]
\centering
\scalebox{.9}
{
\begin{tabular}{c|c|c|c|c|c }
\hline
\textbf{Dataset} & \textbf{ \#Mai } & \textbf{\#Mio} & \textbf{\#Mao} & \textbf{Bs} & \textbf{lr} \\
\hline
\textit{AQuaMuse} & 16,384 & 64  & 256 & 32 & $5\mathrm{e}{-5}$ \\
\textit{CovidET}  & 512   & 25  & 256 & 32 & $5\mathrm{e}{-5}$ \\
\textit{MA-News}  & 2,048  & 64  & 256 & 64 & $5\mathrm{e}{-5}$ \\
\textit{NEWTS}    & 2,048  & 25  & 256 & 32 & $5\mathrm{e}{-5}$ \\
\textit{QMSum}    & 16,384 & 30 & 256 & 32 & $2\mathrm{e}{-5}$ \\
\textit{SQuaLITY} & 16,384 & 256 & 512 & 32 & $1\mathrm{e}{-4}$ \\
\textit{Wikiasp}  & 16,384 & 10 & 256 & 32 & $1\mathrm{e}{-5}$ \\

\hline
\end{tabular}
}
\caption{Finetuning hyper-paramters parameters.  \#Mai, \#Mio, \#Mao, Bs and lr represent Max input length, Min output length, Max output length, batch size and learning rate, respectively.}
\label{tab:hypama_finetune}
\end{table}

\begin{table}[h]
\centering
\scalebox{1.}
{
\begin{tabular}{c|c|c|c }
\hline
\textbf{Dataset} & \textbf{ \#Mai } & \textbf{\#Mio} & \textbf{ \#Mao } \\
\hline
\textit{AQuaMuse}  & 16,384  & 64  & 256 \\
\textit{CovidET}  & 512  & 25   &  256 \\
\textit{MA-News}  & 2,048  & 64  & 256   \\
\textit{NEWTS} &  2,048 &  25  &  256  \\
\textit{QMSum}  & 16,384  & 25   & 256   \\
\textit{SQuaLITY}  & 16,384  & 256   &  512  \\
\textit{Wikiasp}  &  16,384 &  128  & 256   \\

\hline
\end{tabular}
}
\caption{Zero-shot hyper-parameters parameters. \#Mai, \#Mio and \#Mao represent Max input length, Min output length and Max output length, respectively. }
\label{tab:hypama}
\end{table}

\begin{table}[h]
\centering
\scalebox{1.}
{
\begin{tabular}{c|c|c|c }
\hline
\textbf{Dataset} & \textbf{ 0.3 \%} & \textbf{1\%} & \textbf{ 3\% } \\
\hline
\textit{AQuaMuse}  & 17  & 57  & 173 \\
\textit{CovidET}  & 13  & 44   &  132 \\
\textit{MA-News*}  & 85  & 285  & 854   \\
\textit{NEWTS} &  14 &  48  &  144  \\
\textit{QMSum}  & 3  & 12   & 137   \\
\textit{SQuaLITY}  & 3  & 10   &  30  \\

\hline
\end{tabular}
}
\caption{Number of training instances under different few-shot settings. MA-News results are under 0.03\%, 0.1\%, and 0.3\%. }
\label{tab:few-shot-number}
\end{table}

\noindent \textbf{Zero/Few-shot}.
In \cref{tab:hypama}, we show the hyper-parameters used for zero/few-shot settings where no\_repeat\_ngrams is kept at 3/0, the beam size is 4/1, and the length penalty is always set to 1.0. We only use early\_stopping for zero-shot.
The epochs and learning rate for few-shot training are always 60 and 2e-5 respectively. warm-up rates are set to 0.05, while weight decay is 0.01. Batch size is 2 for 0.3\%, while 4 for 1\% and 3\% scenarios.
In \cref{tab:few-shot-number}, we also show the exact number of instances used for few-shot training. The total number of picked training instances ranges from less than ten to several hundred.

\subsection{WikiAsp Full Results}
\label{appendix: wikiasp_full_results}
\begin{table*}[]
\centering
\scalebox{1.}{
\begin{tabular}{l@{\hspace{0.8\tabcolsep}}|c|ccc|ccc}
\hline
\multirow{2}{*}{\textbf{Domain}} & \multirow{2}{*}{\textbf{Models}} & \multicolumn{3}{c|}{\textbf{Finetune }}& \multicolumn{3}{c}{\textbf{Zero-shot }}\\
 \cline{3-8} 
 & & R-1 & R-2 & R-Lsum & R-1 & R-2 & R-Lsum \\
\hline
\multirow{2}{*}{\textbf{Album}} & LED & 19.02 & 7.56 & 17.28 & 7.64 & 0.79 & 6.83 \\
& LED-OASum & \textbf{19.83} & \textbf{7.72} & \textbf{18.04} & \textbf{15.01} & \textbf{2.33} & \textbf{13.17} \\
\hline
\multirow{2}{*}{\textbf{Animal}} & LED & 23.16 & 9.19 & 21.52 & 6.83 & 0.73 & 6.17 \\
& LED-OASum & \textbf{24.16} & \textbf{9.44} & \textbf{22.41} & \textbf{12.97} & \textbf{2.07} & \textbf{11.58} \\
\hline
\multirow{2}{*}{\textbf{Artist}} & LED & 21.12 & 6.76 & 19.42 & 7.56 & 0.89 & 6.91 \\ & LED-OASum & \textbf{21.52} & \textbf{6.77} & \textbf{19.78} & \textbf{14.31} & \textbf{2.17} & \textbf{12.81} \\
\hline
\multirow{2}{*}{\textbf{Building}} & LED & 22.94 & \textbf{7.19} & 21.30 & 13.27 & 1.88 & 12.09 \\
& LED-OASum & \textbf{23.18} & 7.16 & \textbf{21.49} & \textbf{19.75} & \textbf{3.90} & \textbf{17.73} \\
\hline
\multirow{2}{*}{\textbf{Company}} & LED & 18.44 & 4.97 & 16.87 & 9.26 & 1.09 & 8.22 \\
& LED-OASum & \textbf{19.12} & \textbf{5.07} & \textbf{17.50} & \textbf{15.77} & \textbf{2.66} & \textbf{13.99} \\
\hline
\multirow{2}{*}{\textbf{EducationalInstitution}} & LED & 21.12 & 7.46 & 18.96 & 8.72 & 1.17 & 7.66 \\
& LED-OASum & \textbf{21.37} & \textbf{7.85} & \textbf{19.22} & \textbf{15.96} & \textbf{3.05} & \textbf{13.98} \\
\hline
\multirow{2}{*}{\textbf{Event}} & LED & 19.33 & 5.56 & 17.57 & 10.90 & 1.20 & 10.02 \\
& LED-OASum & \textbf{20.62} & \textbf{5.90} & \textbf{18.80}  & \textbf{17.35} & \textbf{3.19} & \textbf{15.65} \\
\hline
\multirow{2}{*}{\textbf{Film}} & LED & 19.53 & 6.77 & 17.85 & 7.45 & 0.78 & 6.84 \\
& LED-OASum & \textbf{20.09} & \textbf{6.98} & \textbf{18.38} & \textbf{15.21} & \textbf{2.80} & \textbf{13.68} \\
\hline
\multirow{2}{*}{\textbf{Group}} & LED & \textbf{18.25} & \textbf{5.21} & \textbf{16.89} & 8.23 & 1.13 & 7.50 \\
& LED-OASum & 18.22 & 4.96 & 16.74 & \textbf{12.84} & \textbf{2.27} & \textbf{11.56} \\
\hline
\multirow{2}{*}{\textbf{HistoricPlace}} & LED & \textbf{27.49} & \textbf{9.96} & \textbf{26.02} & 13.53 & 1.77 & 12.44 \\
& LED-OASum & 26.96 & 9.64 & 25.39 & \textbf{19.34} & \textbf{3.59} & \textbf{17.70} \\
\hline
\multirow{2}{*}{\textbf{Infrastructure}} & LED & 23.68 & 9.72 & 21.96 & 9.49 & 1.06 & 8.44 \\
& LED-OASum & \textbf{23.96} & \textbf{9.75} & \textbf{22.15} & \textbf{14.97} & \textbf{2.34} & \textbf{13.08} \\
\hline
\multirow{2}{*}{\textbf{MeanOfTransportation}} & LED & 22.74 & 7.79 & 20.98 & 9.49 & 1.06 & 8.44 \\
& LED-OASum & \textbf{23.96} & \textbf{8.28} & \textbf{22.11} & \textbf{14.97} & \textbf{2.34} & \textbf{13.08} \\
\hline
\multirow{2}{*}{\textbf{OfficeHolder}} & LED & 24.13 & 9.30 & 22.18 & 8.65 & 1.24 & 7.77 \\
& LED-OASum & \textbf{24.58} & \textbf{9.33} & \textbf{22.61} & \textbf{16.41} & \textbf{3.24} & \textbf{14.42} \\
\hline
\multirow{2}{*}{\textbf{Plant}} & LED & 23.87 & 8.13 & 22.17 & 7.57 & 0.82 & 6.83 \\
& LED-OASum & \textbf{24.59} & \textbf{8.15} & \textbf{22.70} & \textbf{14.61} & \textbf{2.22} & \textbf{13.10} \\
\hline
\multirow{2}{*}{\textbf{Single}} & LED & 20.66 & \textbf{8.04} & 18.95 & 8.39 & 0.99 & 7.52 \\
& LED-OASum & \textbf{21.63} & 7.98 & \textbf{19.68} & \textbf{16.83} & \textbf{3.79} & \textbf{14.85} \\
\hline
\multirow{2}{*}{\textbf{SoccerPlayer}} & LED & 16.93 & 5.52 & 15.10 & 6.79 & 0.83 & 5.98 \\
& LED-OASum & \textbf{17.73} & \textbf{5.88} & \textbf{15.89} & \textbf{10.3} & \textbf{1.49} & \textbf{8.80} \\
\hline
\multirow{2}{*}{\textbf{Software}} & LED & 20.29 & 5.70 & 18.55 & 9.32 & 0.97 & 8.35 \\
& LED-OASum & \textbf{20.67} & \textbf{5.82} & \textbf{18.95} & \textbf{17.99} & \textbf{3.39} & \textbf{16.18} \\
\hline
\multirow{2}{*}{\textbf{TelevisionShow}} & LED & 17.53 & \textbf{4.77} & 15.75 & 7.65 & 0.83 & 6.93 \\
& LED-OASum & \textbf{17.53} & 4.68 & \textbf{15.83} & \textbf{15.02} & \textbf{2.45} & \textbf{13.29} \\
\hline
\multirow{2}{*}{\textbf{Town}} & LED & 43.34 & \textbf{29.37} & 42.33 & 7.86 & 0.85 & 7.27 \\
& LED-OASum & \textbf{43.52} & 29.20 & \textbf{42.34} & \textbf{14.35} & \textbf{2.66} & \textbf{13.00} \\
\hline
\multirow{2}{*}{\textbf{WrittenWork}} & LED & 19.98 & 5.16 & 18.04 & 9.06 & 0.86 & 8.33 \\
& LED-OASum & \textbf{20.45} & \textbf{5.22} & \textbf{18.48} & \textbf{17.14} & \textbf{2.72} & \textbf{15.27} \\
\hline
\end{tabular}
}
\caption{Finetuing and zero-shot performance on Wikiasp datasets.}
\label{tab:wikiasp_full_table}
\end{table*}
In \cref{tab:wikiasp_full_table}, we present all WikiAsp~\citep{hayashi2021wikiasp} 20 domains results with fine-tuning and zero-shot settings. It is obvious that our \textbf{LED-\DATANAME} consistently achieves near-double performance for all domains under almost all ROUGE metrics. The improvements over finetuning results are less substantial but still preserve more than 0.5 points improvements. We attribute this advance comes from the rich knowledge contained in our \DATANAME~ corpus. It is worth noting that the inputs of our \DATANAME~ are close to the outputs of Wikiasp, but we are not sure whether the information seen during our training in encoding has direct help for tuning wikiasp in the decoding stage.

\section{Case Study}
\label{appendix: case_study}

\subsection{\textbf{\DATANAME} Examples}
\label{appendix: oasum_example}
\begin{table*}[]
\centering
\scalebox{0.90}{
\begin{tabular}{ c|c|llll  }
\hline
\textbf{Title}& \textbf{Aspect} & \multicolumn{4}{c}{ \textbf{Summary }} \\ 
\hline
\multirow{4}{*}{\textbf{Pokémon}} & \textit{Name} & \multicolumn{4}{p{0.5\textwidth}}{
{"Pokémon" is a media franchise based on the "Pokémon" video game series created by Satoshi Tajiri and published by Nintendo.} 
}\\
& \textit{Concept} & \multicolumn{4}{p{0.5\textwidth}}{
{Pokémon is a media franchise based on the "Pokémon" video game franchise created by Satoshi Tajiri and published by Nintendo.} 
}\\
& \textit{Criticism and controversy} & \multicolumn{4}{p{0.5\textwidth}}{
{"Pokémon" has been criticized by some fundamentalist Christians over perceived occult and violent themes and the concept of "Pokémon evolution", which they feel goes against the Biblical creation account.} 
}\\
& \textit{In other media} & \multicolumn{4}{p{0.5\textwidth}}{
"Pokémon" has also been the subject of numerous manga series, anime, and video games, as well as the "Pokémon Trading Card Game", a collectible card game.
}\\
& \textit{Cultural influence} & \multicolumn{4}{p{0.5\textwidth}}{\textcolor{red}{
Pokémon is a media franchise based on the video game series of the same name, created by Satoshi Tajiri and published by Nintendo.}
}\\

\hline
\multirow{9}{*}{\textbf{Shanghai}} & \textit{History ; Imperialism} & \multicolumn{4}{p{0.5\textwidth}}{
{Shanghai was founded in 746 as a trading port on the Yangtze River, and became an important sea port during the Ming and Qing dynasties. In the 19th century, international attention to Shanghai grew due to European recognition of its economic and trade potential at the river. In 1842, the Treaty of Nanking opened Shanghai as one of the five treaty ports for international trade.}
}\\
& \textit{History ; Japanese invasion} & \multicolumn{4}{p{0.5\textwidth}}
{
In 1937, the city was occupied by the Japanese during the Battle of Shanghai, which resulted in the occupation of the city by the occupying forces.
}\\
& \textit{Politics ; Administrative divisions} & \multicolumn{4}{p{0.5\textwidth}}
{
Shanghai is the capital and largest city of the People's Republic of China. Shanghai is one of the four municipalities under the direct administration of the Government of China, and is divided into 16 county-level districts.
}\\
& \textit{Economy} & \multicolumn{4}{p{0.5\textwidth}}
{
Shanghai is a global financial center, ranking first in the whole of Asia \& Oceania region and third globally (after New York and London) in the 28th edition of the Global Financial Centres Index, published in September 2020 by Z\/Yen and China Development Institute.
}\\
& \textit{Education and research} & \multicolumn{4}{p{0.5\textwidth}}
{
Shanghai is home to many universities and colleges, including Fudan University, Shanghai Jiao Tong University, Tongji University, East China University of Science and Technology, Donghua University, and Shanghai International Studies University.
}\\
& \textit{Geography } & \multicolumn{4}{p{0.5\textwidth}}
{
Shanghai is located on the east coast of the Yangtze River estuary, which separates it from the provinces of Zhejiang to the south and Jiangsu to the west and north.
}\\
& \textit{Culture ; Arts } & \multicolumn{4}{p{0.5\textwidth}}
{
Shanghai is also known for its art, culture, and architecture, and is home to many museums, including the Shanghai Conservatory of Music, the Shanghai Museum, and the Shanghai Art Museum.
}\\
& \textit{Demographics} & \multicolumn{4}{p{0.5\textwidth}}
{\textcolor{red}{
Shanghai is the second-largest city in mainland China and the capital of the province of Zhejiang.}
}\\
\hline

\end{tabular}
}
\caption{Case study: finetuning results on our \DATANAME~test set.}
\label{tab:case-oasum}
\end{table*}

Here we show two examples of Wikipedia pages \textit{Pokémon}\footnote{\url{https://en.wikipedia.org/wiki/Pok\%C3\%A9mon}} and \textit{Shanghai}\footnote{\url{https://en.wikipedia.org/wiki/Shanghai}} from \DATANAME~ test set in \cref{tab:case-oasum}. The aspect-based summary results are generated by our trained LED-\DATANAME~ checkpoint. It is clear that 4 aspects for \textit{Pokémon} and 7 aspects for \textit{Shanghai} indeed produce strongly relevant and coherent aspect-based summaries. But it still fails for generating correct summaries for aspect \textit{Cultural influence} and \textit{Demographics} highlighted in red. We attribute such errors to coming from two perspectives: the model fails to focus on a certain aspect or it can not generate correct summaries. For example, for \textit{Cultural influence} in \textit{Pokémon}, the generated summary is coherent, fluent, and "correct", but not related to this specific aspect at all. For \textit{Demographics} in \textit{Shanghai}, the first half of the summary is focused on \textit{Demographics}, but the remaining description \textit{the capital of the province of Zhejiang} is both unrelated and inaccurate. 

\subsection{\textbf{LED-\DATANAME} Examples}
\label{appendix: summary_example}
\noindent \textbf{Zero-shot}.
\begin{table*}[t]
    \centering
    \footnotesize
    \resizebox{1.\linewidth}{!}{
            \begin{tabular}{llll}
            \toprule
            Query in AQuaMuse test set: \textit{what do you mean by analysis of variance}\\
            \hline
            \multicolumn{4}{p{1.\linewidth}}{ 
                {
                \textbf{LED-\DATANAME}:
                The analysis of variance (ANOVA) is a collection of statistical tests used to determine whether the variance of a variable in a particular variable is partitioned into components attributable to different sources of variation. For this reason, ANOVA is useful in comparing two, three, or more means. For the purpose of ANOVA, the observed variance in a specific variable (or group of means) is divided by means to determine the probability that the means of several groups are equal, and therefore generalizes the t-test to more than two groups. \textcolor{red}{'rouge1': 64.35, 'rouge2': 42.11, 'rougeL': 42.61, 'rougeLsum': 61.74} }} \\
            \midrule
            \multicolumn{4}{p{1.\linewidth}}{
                {\textbf{LED-base}: This is a summary of the results of the ANOVA. The results of ANOVA are summarized in the following table. This is the summary of all the results from ANOVA: The results are summarized as follows:The ANOVA is a collection of statistical methods used to determine the degree of variance in a given variable. The result is summarized in terms of the number of variance and the types of variance that can be determined. The examples are:The results of AnOVA are summarised as follows. The analysis of variance is summarised in the above table. The example is summarized in the below table.The Example is summarizing the results. The Examples are summarizing all the findings. The Example Is summarizing these results.The Examples are summarized by the following Table.The examples are summarized by the preceding table. They are summarized with the following examples:The Examples of ANOVAs are summarising all the examples. The Methods are summarization of the following:The Example of anOVA is summarising the results and summarizing them in the form of a summary. The method is summarizable by the fact that the results are summarizable. The methods are summarisable by the means of the various methods. The Method is summarization by the form. \textcolor{red}{'rouge1': 26.93, 'rouge2': 8.07, 'rougeL': 16.62, 'rougeLsum': 25.79 }}}\\
            \midrule
            \multicolumn{4}{p{1.\linewidth}}{
                {\textbf{GOLD}: Analysis of variance (ANOVA) is a collection of statistical models and their associated estimation procedures (such as the "variation" among and between groups) used to analyze the differences among group means in a sample. ANOVA was developed by statistician and evolutionary biologist Ronald Fisher. In the ANOVA setting, the observed variance in a particular variable is partitioned into components attributable to different sources of variation. In its simplest form, ANOVA provides a statistical test of whether the population means of several groups are equal, and therefore generalizes the t-test to more than two groups. ANOVA is useful for comparing (testing) three or more group means for statistical significance. It is conceptually similar to multiple two-sample t-tests, but is more conservative, resulting in fewer type I errors, and is therefore suited to a wide range of practical problems.}}\\
            \toprule
            Query in QMSum test set: \textit{summarize the discussion about the vcr / vhr / dvd capability button .}\\
            \hline
            \multicolumn{4}{p{1.\linewidth}}{ 
                {
                \textbf{LED-\DATANAME}:
                The vocalsound button is a button on the vocalsound synthesizer that can be used to control the volume of the synthesizer. The vocalsound can also be used as a remote control. \textcolor{red}{'rouge1': 25.29, 'rouge2': 2.35, 'rougeL': 16.09, 'rougeLsum': 20.69 } }} \\
            \midrule
            \multicolumn{4}{p{1.\linewidth}}{
                {\textbf{LED-base}: summarize the discussion about the vcr / vhr / dvd capability button. marketing: yeah. project manager: okay. marketing : yeah. user interface: yeah, i think it's a good idea to have a button on the tv, i mean it's a button that you can put on your tv, and then you can use the button to put it on your TV, and you can also use the buttons to put the buttons on your television, and if you want to do that, you can do that. marketing, okay. product manager: oh. marketing, okay. user interfaces: yeah. marketing: okay, i'm sure we can do this, we're gonna do it, we're going to do it, we're gonna do this, we 'll do it. we'll do it. we've got a lot of stuff to do, we'll be doing this, and we's going to try to do this. we're going to be doing it, and i 'll be doing the same thing we did last time, we'll do the same, we will do it in the next time. we 'd do it with a button, we could do it on our tv. \textcolor{red}{'rouge1': 17.39, 'rouge2': 3.19, 'rougeL': 10.28, 'rougeLsum': 15.81 }}}\\
            \midrule
            \multicolumn{4}{p{1.\linewidth}}{
                {\textbf{GOLD}: project manager mentioned that the group needed to decide on the vcr capabilities and answered industrial designer 's question on whether vcrs were the same across the world . so industrial designer suggested a whole different set of buttons for individual vcrs . project manager suggested dvd type if considering the target group as young professionals and teenagers .}}\\

            \toprule
            Aspect in NEWTS test set: \textit{attack, attacks, killed, attacked, bomb, people, explosion, bombing, injured, blast,Cairo}\\
            \hline
            \multicolumn{4}{p{1.\linewidth}}{ 
                {
                \textbf{LED-\DATANAME}:
                \textcolor{green}{At least 12 people were killed and more than 40 were injured in the attacks, according to the Egyptian government. The attacks occurred during the ongoing insurgency in Egypt's northern Sinai Peninsula.} \textcolor{red}{'rouge1': 23.91, 'rouge2': 8.89, 'rougeL': 15.22, 'rougeLsum': 19.57 } }} \\
            \midrule
            \multicolumn{4}{p{1.\linewidth}}{
                {\textbf{LED-base}: attack, attacks, killed, attacked, bomb, people, explosion, bombing, injured, blast,Cairo (CNN)At least 12 people were killed Sunday, and more injured, in separate attacks on a police station, a checkpoint and along a highway in Egypt's northern Sinai, authorities said. Six people, including one civilian, were killed when a car bomb exploded near the police station in Al-Arish, capital of North Sinai, Health Ministry spokesman Hossam Abdel-Ghafar told Ahram Online. He said 40 people were injured. Ansar Beit Al-Maqdis, an ISIS affiliate, claimed responsibility for the attack, which came hours after another operation that the group also claimed. In that earlier attack, a first lieutenant, a sergeant and four conscripts were killed in an armored vehicle. Two other soldiers were injured and taken to a military hospital. Another soldier was killed in the second attack, the military said in a statement. The military said the attack was carried out by an unknown assailant, who was not identified by name, but was believed to be a member of the Islamic State of Iraq and the Levant (ISIL), a militant group affiliated with al-Qaida in the Islamic Maghreb. The army said the. \textcolor{red}{'rouge1': 36.08, 'rouge2': 20.55, 'rougeL': 25.88, 'rougeLsum': 32.94 }}}\\
            \midrule
            \multicolumn{4}{p{1.\linewidth}}{
                {\textbf{GOLD}: Twelve people in total were killed at a police station, a checkpoint, and along a highway in Egypt. Six people were killed in a car bomb near the police station and forty were injured. Six other people were killed when their armored car was attacked and two were injured. Three security personnel were also injured after being shot at.}}\\
                
            \bottomrule
            \end{tabular}
    }
\caption{Examples of aspect-based summaries under zero-shot setting.}
\label{tab:case-led-oasum}
\end{table*}

Here we show three examples from downstream AQuaMuse, QMSum, and NEWTS datasets under the zero-shot setting in \cref{tab:case-led-oasum}. As we can see from the results of AQuaMuse and QMSum, LED-\DATANAME~ can produce much better summaries. For another example from NEWTS, although LED-base achieves higher rouge scores, the summary is actually redundant and repetitive. On the contrary, the LED-\DATANAME~ generated summary(highlighted in \textcolor{green}{green}) preserves the summary towards the chosen aspect and demonstrates good quality.

\noindent \textbf{Few-shot}.
\begin{table*}[t]
    \centering
    \footnotesize
    \resizebox{1.\linewidth}{!}{
            \begin{tabular}{llll}
            \toprule
            Zero-shot: Query in SQuALITY test set: \textit{What is the significance of Ghost Ships in the story?}\\
            \hline
            \multicolumn{4}{p{1.\linewidth}}{ 
                {
                \textbf{LED-\DATANAME}:
               Space Man Willard is a short story by American writer James Willard, published in 1942. The story is about a space ship that Willard had never seen before. The ship was lost for twenty years, and Willard was the only survivor of the Mary Lou, which had been destroyed by a meteor. Willard's ship, the "Mary Lou II", was the first space ship to sail beyond the Solar System, and the only ship to have been lost for over a thousand years. The Mary Lou was named in honor of his father, who died when the ship was still in space. It was the last ship Willard would see before his ship would be able to return to Earth. The ghost of the ship would come back to him, and he would never see it again. It would be the last time he would see Earth again, and it would be too late for him to see the ship again. He would spend the next twenty years of his life in space alone, and would never return to the Earth again. In the story, Willard imagines that the ghost ship will come to Earth and bring him home. He imagines the ship will be the home of spacemen who can never go home again. And this is what happened to. \textcolor{red}{'rouge1': 45.78, 'rouge2': 10.71, 'rougeL': 16.32, 'rougeLsum': 35.52} }} \\
            \midrule
            \multicolumn{4}{p{1.\linewidth}}{
                {\textbf{LED-base}: What is the significance of Ghost Ships in the story? Willard and Larry Dobbin left the Rocket Port for the exploration of an alleged planetoid beyond Pluto. The ship has not been seen or heard from since. J. Willard, II, son of the lost explorer, is planning the manufacture of a super-size exploration ship to be called Mary Lou II, in memory of his father." Chorus TextColor TextColor He awoke with a start and a cry of alarm ran through him as he thought that perhaps he might still be in the Mary Lou. The warm, smiling face of a man quickly reassured him.Chorus FontSize TextColor "I'll call the captain," the space man said. "He said to let him know when you came to."Chorus of TextColor The ghost ship was a ghost ship. It was the ghost ship, the ghost of a ghost. The ghost ships were ghost ships, ghost ships and ghost ships. The Ghost Ship was ghost ships of ghost ships in the Ghost Ship in his youth.chorus escription TextColor It was ghost ship of ghost ship TextColor When Willard awoke, he could not sleep. And so it was that each night—for Willard did not give up the Earth-habit of keeping time—Willard dreamed of the days he had known on Earth. chorus TextColor A ghost ship in the sky. "Nonsense!" Willard broke in hurriedly, hoping that the dying man would not see through the lie. "We've got the sun's gravity helping us drift back to Earth! We'll be there soon! You'll get well soon and we'll start to work again on a new idea of mine...." His voice trailed helplessly away and the words were lost. He was no longer able to sleep. Chorus of the Ghost ship in space TextColor Chorus Of the ghost ships In the sky, he thought, "It's all right," Willard whispered. The sick man did not hear him. Two tears rolled down his cheeks. His face contorted as he tried to withhold a sob.Chrome TextColor ITextColor "How do you feel, Space Man Willard?" Chorus, chorus of ghosts TextColor chorus chorus "Oh, you know me?" Willard looked at him in surprise, and then smiled, "I don't know you."chorus Of TextColor Darkness TextColor There was nothing to see. Darkness - Darkness  Archdemon Darkness  Gleaming Darkness \textcolor{red}{'rouge1': 37.12, 'rouge2': 9.31, 'rougeL': 16.32, 'rougeLsum': 35.52 }}}\\
            \midrule
            \multicolumn{4}{p{1.\linewidth}}{
                {\textbf{GOLD}: The Ghost Ship is a legend that sailors and space travelers alike have claimed people see in the moments before they die at sea or in space. In the seconds before Dobbin dies in Willard’s arms, he looks out the window of the “Mary Lou” and claims to see the Ghost Ship himself. Throughout Willard’s long periods of solitude aboard the “Mary Lou”, he thinks he sees the Ghost Ship several times. First, from a distance, as a blinking light advancing closer and closer before turning back and sailing off into dark space; later, he thinks he sees the ship return, only this time it passes nearer before turning back and leaving again. With each return of the Ghost Ship, Willard believes he sees it clearer than he had before. After decades adrift in space, Willard believes a ship has finally come to rescue him. He does not think it is the Ghost Ship because it is solid, and he is greeted by a crew of people. However, the captain explains that the longer a vessel spends lost in space, the more it loses itself and slips into a kind of un-reality, along with those aboard. The more the “Mary Lou” drifted into this space, the more real the Ghost Ship became to Willard. Willard realizes that the “Mary Lou” has become a “ghost ship” herself.}}\\
            \toprule
            Few-shot: Query in SQuALITY test set: \textit{What is the significance of Ghost Ships in the story?}\\
            \hline
            \multicolumn{4}{p{1.\linewidth}}{ 
                {
                \textbf{LED-\DATANAME}:
                Ghost Ships are ghost-like ships that haunt the Solar System that gave it birth. They are the home of spacemen who can never go home again. When your last measure of fuel is burnt and your ship becomes a lifeless hulk—the Ghost will come—for you. Space Man Willard has a friend who died in space, J. Willard II, the son of the lost explorer. He is planning to manufacture a super-size exploration ship to be called Mary Lou II, in memory of his father. Each night he sees the sight of the ship and thinks he will see it again. However, after several dreams and visions of the ghost ship haunting him, it is clear that the ship is not a phantom ship. Instead, a life and blood rocket ship from Earth is tormenting him once again. Starlight shines on it and not through it! Its lines, window, vents were all solid and had none of the spirit-like quality he remembered in the Ghost Ship in his youth. He yawns and sighs as he remembers the days he had spent on Earth. Although Willard believed that the ghost ships were superstitions, he saw them from his ship, and he tried to rip the memories of Earth out. \textcolor{red}{'rouge1': 45.80, 'rouge2': 10.48, 'rougeL': 21.31, 'rougeLsum': 42.63 } }} \\
            \midrule
            \multicolumn{4}{p{1.\linewidth}}{
                {\textbf{LED-base}: What is the significance of Ghost Ships in the story?GALACTIC GHOSTS IN SPACEWillard is the captain of the Mary Lou ship. He is also a friend of Dobbin, a scientist who had come to explore the Solar System. He knows that Dobbin has died a horrible death, and the stories that he tells about the ghost ship that he has read about in the newspaper. He also knows that the unnamed unnamed ship which came to capture Dobbin’s attention is a ghost ship. However, despite these beliefs, Willard does not believe that the ship has actually landed on the planet and is actually traveling to see Dobbin again. For a moment the reason in his mind tottered on a balance. Doubt assailed him. Was this the Ghost Ship come to torment him again? But no, it does not. He is greeted by a small crowd of people and his room is filled with empty rooms. The ghost ship seems to have abandoned its passengers and crew members, leaving Willard alone in his room. He does not understand that his friends have just landed their ship here and are experiencing the land for the first time.  perspectemaleHe is taken aback by the large crowd that has gathered around him, and he begins to imagine that his ship has landed on Earth. He would see Earth again! That single thought runs through his mind constantly. The tapping of the space-telegrapher reassured him. He hears a message from the ship that says "CALLING SPACE SHIP MARY LOU," the message rapped out, "Yes, that is it!” With trembling fingers that he could scarcely control, old Willard sent the answering message. It is considered to be the most important message of the story. \textcolor{red}{'rouge1': 44.23, 'rouge2': 13.13, 'rougeL': 20.77, 'rougeLsum': 41.93 }}}\\

            \bottomrule
            \end{tabular}
    }
\caption{Examples of aspect-based summaries under zero/few-shot setting. Few-shot means the model is finetuned on randomly chosen 3\% samples from the training set.}
\label{tab:case-led-oasum2}
\end{table*}

Besides, we also show one example from SQuALITY dataset under few/zero-shot setting in \cref{tab:case-led-oasum2}. Under the zero-shot conditions, our LED-\DATANAME~ can generate a much better query-based summary than the original LED-base model, which can also be observed from ROUGE scores. When the models are furthermore tuned on a small amount of 3\% (30) of training instances, the improvements mainly come from ROUGE-L and ROUGE-LSum. 

\end{document}